\pdfoutput=1

\documentclass[11pt,a4paper]{article}

\usepackage[acceptedWithA]{tacl2018v2}
\usepackage{tikz}

\usepackage{times}
\usepackage{latexsym}
\usepackage{url}

\usepackage[T1]{fontenc}

\usepackage[utf8]{inputenc}

\usepackage{microtype}

\usepackage{times,latexsym}
\usepackage{url}
\usepackage[T1]{fontenc}
\usepackage{booktabs} 
\usepackage{adjustbox}
\usepackage{CJKutf8}
\usepackage{comment}

\newcommand*\circled[1]{\tikz[baseline=(char.base)]{
    \node[shape=circle,draw,inner sep=0.5pt] (char) {#1};}}
\newcommand{\abr}[1]{\textsc{#1}}
\newcommand{\camelabr}[2]{{\textsc{#1}}{\textsc{#2}}}
\newcommand{\shuowen}{\camelabr{Shuo}{Wen}}

\newcommand{\plm}[0]{{PLM}}
\newcommand{\cws}[0]{\abr{cws}}

\newcommand{\ulm}[0]{\abr{ulm}}

\title{Sub-Character Tokenization for Chinese Pretrained Language Models}

\author{Chenglei Si$^{1,2*}$, Zhengyan Zhang$^{1*}$, Yingfa Chen$^{1*}$, Fanchao Qi$^{1}$,  \\
 \textbf{Xiaozhi Wang}$^{1}$, \textbf{Zhiyuan Liu}$^{1\dagger}$, \textbf{Yasheng Wang}$^3$, \textbf{Qun Liu}$^3$, \textbf{Maosong Sun}$^{1\dagger}$\\
  $^1$ NLP Group, DCST, IAI, BNRIST, Tsinghua University, Beijing, China \\
  \texttt{\{zy-z19,yingfa-c18,qfc17,wangxz20\}@mails.tsinghua.edu.cn} \\
  \texttt{\{liuzy,sms\}@tsinghua.edu.cn}\\
  $^2$ University of Maryland, College Park, MD, USA \\
  \texttt{clsi@terpmail.umd.edu} \\
  $^3$Huawei Noah's Ark Lab, Hong Kong, China \\
         \texttt{\{wangyasheng,qun.liu\}@huawei.com}\\
}

\date{}

\begin{document}
\maketitle
\begin{CJK*}{UTF8}{gbsn}
\begin{abstract}
Tokenization is fundamental to pretrained language models (\plm{}s).
Existing tokenization methods for Chinese \plm{}s typically treat each character as an indivisible token. However, they ignore the unique feature of the Chinese writing system where additional linguistic information exists below the character level, \textit{i.e.}, at the sub-character level.
To utilize such information, we propose sub-character (SubChar for short) tokenization.
Specifically, we first encode the input text by converting each Chinese character into a short sequence based on its glyph or pronunciation, and then construct the vocabulary based on the encoded text with sub-word segmentation. 
%
%
Experimental results show that SubChar tokenizers have two main advantages over existing tokenizers: 1) They can tokenize inputs into much shorter sequences, thus improving the computational efficiency. 2) Pronunciation-based SubChar tokenizers can encode Chinese homophones into the same transliteration sequences and produce the same tokenization output, hence being robust to homophone typos.
At the same time, models trained with SubChar tokenizers perform competitively on downstream tasks. 
We release our code and models at \url{https://github.com/thunlp/SubCharTokenization} to facilitate future work.

\end{abstract}

\section{Introduction}

{\let\thefootnote\relax\footnotetext{$^*$ Equal contribution}}
{\let\thefootnote\relax\footnotetext{$^\dagger$ Corresponding authors}}

\begin{figure*}[t]
\centering
\includegraphics[trim={0.5cm 4.5cm 0.5cm 1.5cm},width=\textwidth]{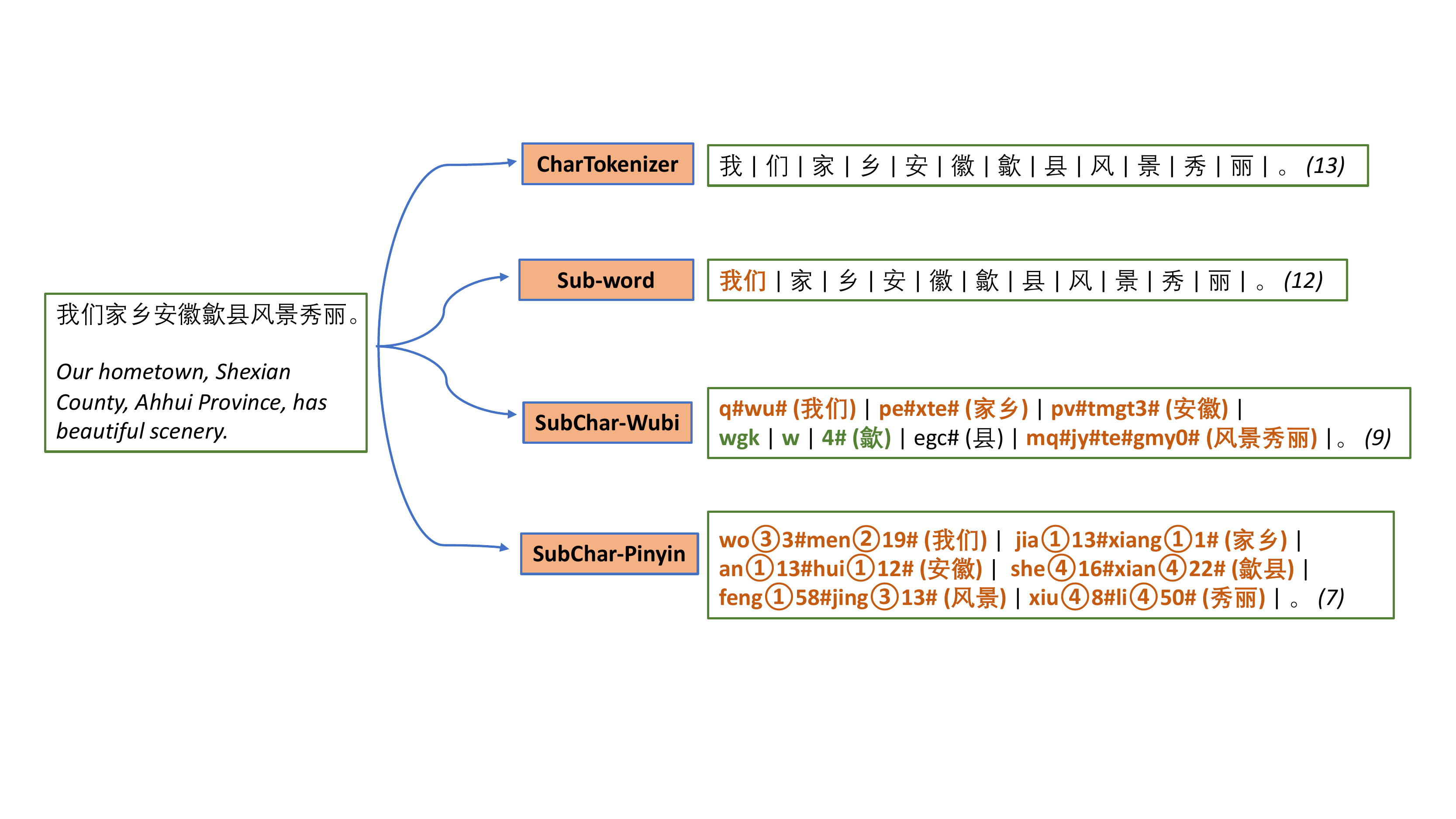}
\caption{Comparison of existing tokenizers (character tokenizer and sub-word tokenizer) and our sub-character tokenizers (SubChar-Wubi using glyph and SubChar-Pinyin using pronunciation encoding).
Different tokens produced by the tokenizers are separated by `|'. The numbers in \textit{(brackets)} indicate the number of tokens in the tokenized sequence.
Tokens in \textcolor{BrickRed}{orange} indicate character combinations, while tokens in \textcolor{OliveGreen}{green} indicate sub-character tokens. `\#' indicates the special separation symbol after each character, circled numbers (\circled{1}\circled{2}\circled{3}\circled{4}) indicate the intonation of characters.  (Figure best viewed in color.)}
\end{figure*}



Large-scale Transformer-based pretrained language models (\plm s)~\cite[][\textit{inter alia}]{BERT,RoBERTa,ALBERT,ELECTRA,DeBERTa} have achieved great success in recent years and attracted wide research interest, in which tokenization plays a fundamental role.

The most popular type of tokenization adopted by \plm{}s is sub-word tokenization, such as byte pair encoding (BPE)~\cite{BPE}, WordPiece~\cite{WordPiece} and unigram language model segmentation~\cite{Kudo2018}. 
Recent Chinese \plm{}s such as CPM~\cite{CPM,CPM2} adopt this kind of sub-word tokenization.
Apart from sub-word tokenization, many other Chinese \plm{}s adopt a simple character tokenizer (CharTokenizer for short) that treats every single Chinese character as a token~\cite[][\textit{inter alia}]{ERNIE,WWM,MacBERT}.


However, we believe that both of these existing tokenizers are sub-optimal for Chinese. This is based on the observation that Chinese has unique linguistic characteristics:
%

1) Chinese has an opaque orthography with irregular grapheme-phoneme correspondence~\cite{phonology_new}. This is in contrast to transparent orthographies like Spanish and Finnish where each letter approximately represents one sound.
As a result, utilizing pronunciation information in Chinese requires explicit pronunciation encoding.
%

2) Chinese does not have morphological inflection, unlike morphologically-rich languages like Russian~\cite{writing_system}. This renders sub-word tokenization less useful since the main advantage of sub-word tokenization comes from the fact that it can split common affixes and root words as separate tokens. In fact, Chinese characters are logograms, and their glyphs (the composition of radicals) also contain rich semantic information, which can only be captured at the sub-character level.


Motivated by these observations, we propose the novel sub-character (SubChar) tokenization. It first encodes every Chinese character into a sequence of phonetic or stroke symbols, and then it uses a sub-word segmenter (such as BPE) to construct the vocabulary on all the encoded sequences. 
In this way, the resultant tokenizers can capture sub-character tokens that correspond to meaningful phonetic or morphemic units, which are absent from all existing Chinese tokenizers. As far as we know, this is the first attempt on leveraging the sub-character information for language models, especially in the context of Chinese NLP.

To assess the effectiveness of our proposed method, we train a series of BERT-style \plm s using the existing and proposed tokenizers.
We evaluate these models on over ten datasets of various downstream natural language understanding (NLU) tasks. 
Through extensive evaluation, we find that models trained with SubChar tokenizers match models trained with character and sub-word tokenizers on downstream task performance. 
More importantly, SubChar tokenizers have two major advantages compared to existing tokenizers:

  1) \textbf{SubChar tokenizers are more efficient.} We find that a small fraction of sub-character tokens in the vocabulary can compose a large variety of rare and complex characters, thus saving much space in the vocabulary for more character combination tokens such as words and phrases. The increased use of combination tokens leads to significantly decreased length of the tokenized sequences. For example, on the iFLYTEK long text classification dataset, with the same vocabulary size as the CharTokenizers, SubChar tokenizers can achieve as much as 40\% length reduction on the tokenized output. Such length reduction can significantly speed up both pretraining and finetuning.

 2) \textbf{SubChar tokenizers are more robust.} A common and unique type of typos in Chinese is caused by homophones where characters with different semantic meanings have exactly the same pronunciation. SubChar tokenizers based on pronunciation can map homophones into the same transliteration sequences, thus improving robustness against any homophone typos. This could be immensely useful when handling noisy inputs. 




We believe that our work is an important step towards more tailored techniques for languages beyond just English by effectively integrating the unique linguistic characteristics of the language~\cite[\#BenderRule]{BenderRule}. 


\section{Method}

\label{sec:method}

In this section, we describe our proposed SubChar tokenization in detail. We break it down into two steps: 1) Chinese character encoding; 2) vocabulary construction based on the encoded sequences. 


\subsection{Step 1: Character Encoding}
\label{sec:char_encode}

The core idea of this step is to encode every Chinese character into a sequence that characterizes its glyph or pronunciation, in order to provide additional inductive biases to the model. We explore several ways of encoding the characters.
They can be categorised as pronunciation-based and glyph-based encoding.

\paragraph{Pronunciation-based encoding}
In order to capture pronunciation information of characters, we encode Chinese characters using transliteration, which uses IPA-inspired\footnote{IPA: International Phonetic Alphabet (\url{https://en.wikipedia.org/wiki/International_Phonetic_Alphabet})} phonetic scripts to characterize the pronunciation.

We explore two different transliteration methods: \textit{pinyin} and \textit{zhuyin (i.e., bopomofo)}.  
\textit{Pinyin} uses romanized transcription and four different tones (\={}, \'{}, \v{}, \`{})
to transliterate characters, \textit{e.g., 魑魅魍魉} $\rightarrow$ \textit{Chi\={} Mei\`{} Wang\v{} Liang\v{}}. 
On the other hand, \textit{zhuyin} uses a set of graphemes nonexistent in English and the same four tones to transliterate the characters, \textit{e.g., 魑魅魍魉} $\rightarrow$ \textit{ㄔ ㄇㄟ\`{} ㄨㄤ\v{} ㄌㄧㄤ\v{}}. 
In \textit{zhuyin}, the first tone mark (\={}) is usually omitted.

We insert special separation symbols (\textit{\#}) after each character's transliterated sequence, \textit{e.g., Chi\={}\#Mei\`{}\#Wang\v{}\#Liang\v{}\#, ㄔ\#ㄇㄟ\`{}\#ㄨㄤ\v{}\#ㄌㄧㄤ\v{}\#}. This prevents cases where transliterated sequences of different characters are mixed together, especially when there are no tone markers to split them in \textit{zhuyin}.
  
 Different Chinese characters may have the same pronunciation even if they have different semantic meanings (\textit{i.e.,} homophones). For disambiguation, we append different indices after the encoded sequences for the homophonic characters, so as to allow a biunique mapping between each Chinese character and its transliteration sequence, \textit{e.g., Chi\={}33\#Mei\`{}24\#Wang\v{}25\#Liang\v{}13\#, ㄔ10\#ㄇㄟ\`{}3\#ㄨㄤ\v{}6\#ㄌㄧㄤ\v{}1\#}. 

It is unclear whether having such disambiguation of homophones is beneficial or not. To analyse the impact, we also experiment with a variant where we do not add the indices to disambiguate the homophones. We implement the tokenizer \textit{SubChar-Pinyin-NoIndex} to perform \textit{pinyin} encoding without disambiguation indices.
We will show that this variant also has the advantage of being robust to homophone typos~(section \ref{sec:noisy_eval}). 

\paragraph{Glyph-based encoding}
The glyphs (\textit{i.e.}, shapes) of Chinese characters contain rich semantic information and can help NLP models~\cite{cw2vec}.  Most Chinese characters can be broken down into semantically meaningful radicals. Characters that share common radicals often have related semantic information, e.g., the four characters \textit{`魑魅魍魉'} share the same radical \textit{`鬼'} (meaning ``ghost''), and their meanings are indeed all related to ghosts and monsters.\footnote{The word \textit{`魑魅魍魉'} is in fact a Chinese idiom, which is often used to refer to bad people who are like monsters.} 
In order to capture glyph information, we explore four glyph-based encoding methods, namely \textit{Stroke, Wubi, Zhengma}, and \textit{Cangjie}.






For \textit{stroke} encoding, we use the Latin alphabet to represent the set of Chinese strokes and convert the characters based on the standard stroke orders,\footnote{\url{https://en.wikipedia.org/wiki/Stroke_order}} \textit{e.g., 魑} $\rightarrow$ \textit{\underline{pszhshpzzn}nhpnzsszshn}; \textit{魅} $\rightarrow$ \textit{\underline{pszhshpzzn}hhspn} (underlined parts indicate shared stroke sequences across these characters). 

The other three glyph-based encoding methods encode characters into radical sequences instead,
by using glyph-based Chinese input methods: \textit{Wubi}, \textit{Zhengma} and \textit{Cangjie}. These input methods group strokes together in different ways to form radicals, and then decompose characters into radical sequences. We use the Latin alphabet to represent these radicals, \textit{e.g., 魑魅魍魉} $\rightarrow$ \textit{Wubi: \underline{rqc}c \underline{rqc}i \underline{rqc}n \underline{rqc}w}; \textit{Zhengma: \underline{nj}lz \underline{nj}bk \underline{nj}ld \underline{nj}oo}; \textit{Cangjie: \underline{hi}yub \underline{hi}jd \underline{hi}btv \underline{hi}mob} (\underline{underlined} parts indicate common radicals among them).

We append the same separation symbol (`\#') after each character, and also add the disambiguation indices for characters whose stroke sequences are identical (\textit{e.g., 人} (people) and \textit{八} (eight)). However, we note that there are very few cases where different characters have the same glyph encoding.


\subsection{Step 2: Vocabulary construction}

Once we have the encoded sequences, we can treat the encoding of each character as the equivalent of `word' in English and then apply sub-word segmentation to construct the vocabulary for our sub-character tokenizers. 

Sub-word segmentation typically forms sub-word tokens by merging frequent token bigrams, which often results in meaningful morphemes of the words when used in languages like English. On our encoded sequences, sub-word segmentation can capture shared sub-character sequences that correspond to shared radicals or phonetic sequences among similar characters. 
After running the sub-word segmentation step on the encoded sequences, the vocabulary of the resultant sub-character tokenizers consists of a mixture of sub-character tokens, character tokens, and character combination tokens. 

In this work, we use the unigram language model segmentation method~\cite{Kudo2018} implemented in SentencePiece~\cite{sentencepiece} as the default sub-word segmentation method. In section~\ref{sec:bpe}, we also perform an ablation study by setting the sub-word segmentation method to BPE, which results in similar performance and efficiency, illustrating that the gains of SubChar tokenization are insensitive to the specific choice of sub-word segmentation methods.

\begin{figure}[t]
\centering
\includegraphics[width=0.5\textwidth]{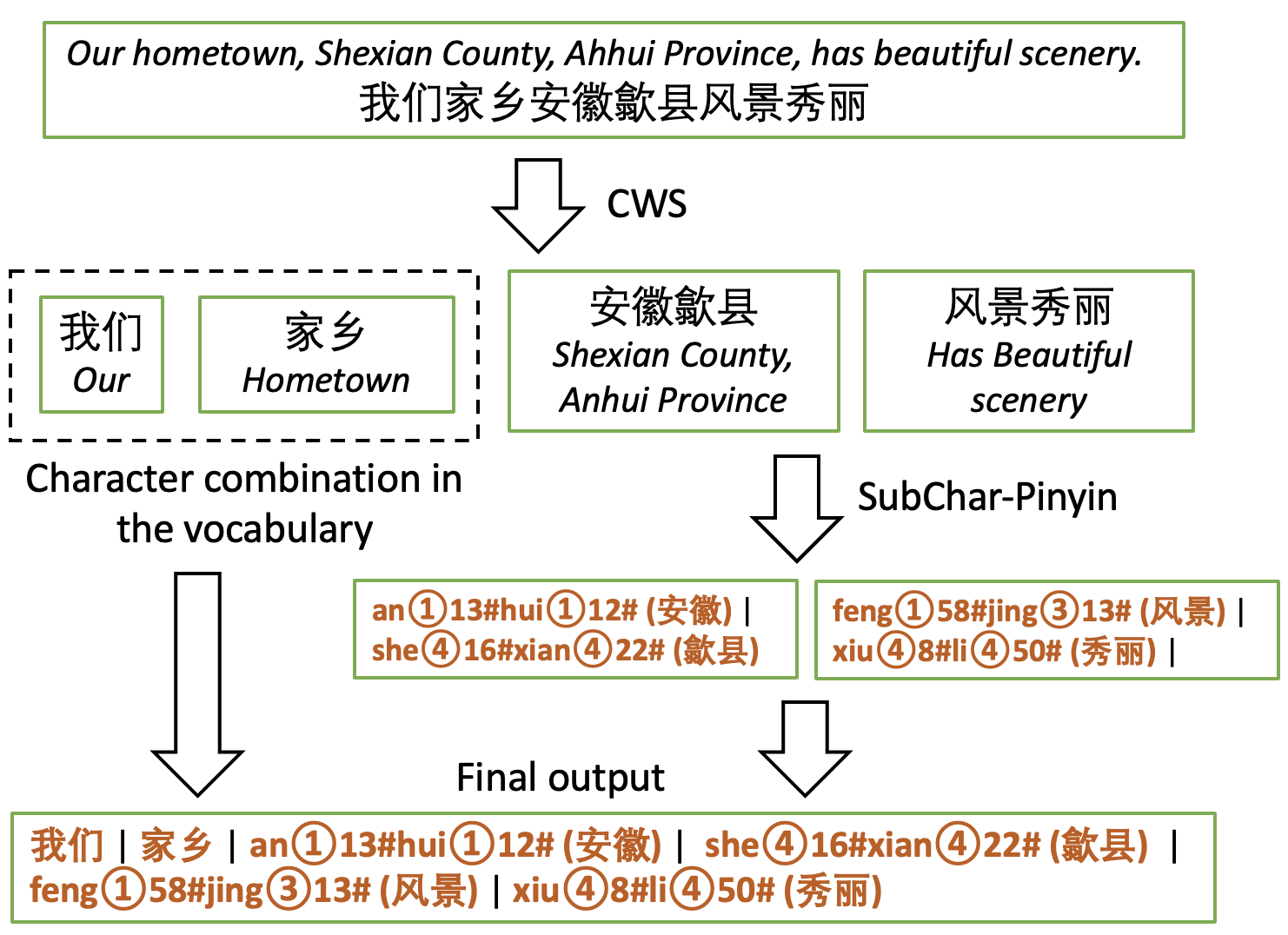}
\caption{Illustration of the tokenization pipeline when incorporating CWS. After the first step of CWS, high-frequency words (words in the dashed box) directly become part of the final output sequence, the other words then go through SubChar tokenization.}
\label{fig:cws}
\end{figure}

\subsection{Optional Step: Chinese Word Segmentation}
\label{sec:cws}

Before the first step of character encoding, there is an optional step of Chinese word segmentation. 

Chinese word segmentation (\cws{}) is a common technique to split Chinese text chunks into a sequence of Chinese words. The resultant segmented words sometimes provide better granularity for downstream tasks~\cite{CWS-NMT}. However, the impact of \cws{} is unclear in the context of pretraining, especially its interplay with the tokenization. Hence, we propose a way to incorporate \cws{} into our SubChar tokenization and examine whether it is helpful. Our proposed tokenization pipeline is summarized in Figure~\ref{fig:cws}.

Given that the vocabulary of SubChar tokenizers consists of character combinations, characters, and sub-characters, we use \cws{} to construct the character combination part of the vocabulary. Compared to the character combination tokens generated by the statistical approach of sub-word tokenization, the combination tokens generated by a trained Chinese word segmenter have more linguistic prior knowledge. 

Specifically, to construct the vocabulary, we first segment the pretraining corpus into words. Then, we select the most frequent words as the character combination part of the SubChar tokenizer vocabulary. We then encode the text with one of the pronunciation- or glyph-based encoding methods and use sub-word tokenization on the encoded sequences to get the sub-character and character tokens of the vocabulary.  Finally, we merge these parts together as the vocabulary for the SubChar tokenizer. When tokenizing new inputs, we first segment them into words, if the words are in the vocabulary, they will be tokenized as word tokens; if not, they will be further processed by the SubChar tokenizer.
We control the ratio of word tokens in the vocabulary to be 80\% based on preliminary tuning and we use a state-of-the-art segmenter THULAC~\cite{THULAC,THULAC-repo} for word segmentation.



  


\section{Experiment Setup}


\begin{table}[t]
    \centering
    \small {
    \begin{tabular}{ lrrr }
        \toprule Dataset & \#Train & \#Dev & \#Test \\
        \midrule
            TNEWS  &  53.4K  & 10K   & 10K   \\
            IFLYTEK  & 12.1K  & 2.6K  & 2.6K   \\
            BQ  &  100K  & 10K   & 10K   \\
            THUCNEWS  & 669K   & 83.6K & 83.6K  \\
            CLUEWSC  &  1.2K  & 0.3K  & 0.3K  \\
            AFQMC    & 34.3K & 4.3K & 3.9K    \\
            CSL &  20K  & 3K   & 3K  \\
            OCNLI   & 45.4K  & 5K   & 3K    \\
            CHID    & 519K  & 57.8K & 23K    \\
            C3      & 12K   & 3.8K  & 3.9K   \\
            CMRC  & 10K   & 3.4K  & 4.9K   \\
            CLUENER  &  11K  & 1.3K  & 1.3K  \\
        \bottomrule
    \end{tabular}}
    \caption{Statistics of downstream datasets.} 
    \label{tab:hyper_parameters}
\end{table}

In this section, we introduce our baselines, datasets and experiment settings.




\subsection{Baselines}

We compare two existing tokenization methods as baselines, namely single-character tokenization and sub-word tokenization.  
For a fair comparison, we set the same vocabulary size of $22,675$ for all tokenizers, including baselines and our proposed tokenizers. This is consistent with the vocabulary size of Chinese BERT~\cite{BERT}.

\subsection{Pretraining Data}


We use the same training corpus to train all the tokenizers in this work. 
The corpus consists of $2.3$ GB Chinese text from Baidu Baike.\footnote{\url{https://baike.baidu.com/}}

To evaluate the effectiveness of the tokenizers, we pretrain a BERT\footnote{Note that we mean BERT-style pretrained Transformers. Our models are not directly comparable with the original Chinese BERT since we use different pretraining data and hyper-parameters.} model using each tokenizer and compare their performance on downstream tasks. 
When pretraining the BERT models, we use the same pretraining corpus (\textit{i.e., } Baidu Baike) and the same set of hyper-parameters.
Notably, we also pretrain a new BERT model using the character tokenizer on our pretraining corpus instead of loading from existing checkpoints~\cite{BERT} so that it provides an apple-to-apple comparison with our proposed methods.
Since our proposed tokenizers are direct drop-in replacements for the baseline tokenizers, they do not incur any extra parameters. 
In summary, all the compared models have the same training corpus, hyper-parameters, and number of parameters, allowing for a truly fair comparison.

\subsection{Evaluation Data}


We finetune and evaluate the pretrained models with different tokenization methods on various downstream NLU datasets, including single-sentence classification, sentence-pair classification, and reading comprehension tasks. We briefly introduce each dataset below and present the dataset statistics in Table~\ref{tab:hyper_parameters}.

\noindent \textbf{TNEWS}~\cite{CLUEBenchmark} is a news title classification dataset containing 15 classes. 

\noindent \textbf{IFLYTEK}~\cite{CLUEBenchmark} is a long text classification dataset containing 119 classes. The task is to classify mobile applications into corresponding categories given their description.

\noindent \textbf{BQ}~\cite{BQ} is a sentence-pair question matching dataset extracted from an online bank customer service log. The goal is to evaluate whether two questions are semantically equivalent.

\noindent \textbf{THUCNEWS}~\cite{THUCNEWS} is a document classification dataset with 14 classes. The task is to classify news into the corresponding categories given their title and content.

\noindent \textbf{CLUEWSC}~\cite{CLUEBenchmark} is a coreference resolution dataset in the format of Winograd Schema Challenge~\cite{levesque2012winograd}. The task is to determine whether the given noun and pronoun in the sentence refer to the same entity.

\noindent \textbf{AFQMC}~\cite{CLUEBenchmark} is the Ant Financial Question Matching Corpus for the question matching task that aims to predict whether two sentences are semantically equivalent.

\noindent \textbf{CSL}\footnote{\url{https://github.com/P01son6415/CSL}} is the Chinese Scientific Literature dataset extracted from academic papers. Given an abstract and some keywords, the goal is to determine whether they belong to the same paper. It is formatted as a sentence-pair classification task.

\noindent \textbf{OCNLI}~\cite{OCNLI} is a natural language inference dataset. The task is to determine whether the relationship between the hypothesis and premise is entailment, neutral, or contradiction.

\noindent \textbf{CHID}~\cite{CHID} is a  cloze-style multiple-choice reading comprehension dataset. Given a context where some idioms are masked, the task is to select the appropriate idiom from a list of candidates.

\noindent \textbf{C3}~\cite{C3} is a multiple-choice reading comprehension dataset. The goal is to choose the correct answer for the questions given context. 

\noindent \textbf{CMRC}~\cite{CMRC} is a span-extraction reading comprehension dataset consisting of questions annotated from Wikipedia paragraphs.

\noindent \textbf{CLUENER2020}~\cite{CLUENER} is a named entity recognition dataset with 10 entity types.

\subsection{Hyper-parameters}

We elaborate on all hyper-parameters involved for reproducibility (we also release all code, trained tokenizers and models).

\textbf{Tokenizer Training}.
When training tokenizers with SentencePiece, we use a character coverage of $1.0$ and model type `unigram' for all tokenizers being compared. Other hyper-parameters follow the default of SentencePiece.

\begin{table*}[t]
    \centering
    \small
    \begin{adjustbox}{width=\linewidth,center}
    \setlength{\tabcolsep}{3.5pt}{
    \begin{tabular}{ lccccccccccc }
    \toprule 
     & TNEWS
     & IFLY
     & THUC
     & BQ 
     & WSC
     & AFQMC
     & CSL
     & OCNLI
     & CHID
     & C3
     & AVG
     \\
     \hline  
    \multicolumn{12}{c}{\textit{6-layer, 2.3G Corpus}} \\
    \hline 
CharTokenizer
& \begin{tabular}{@{}c@{}}{64.19} \\ \footnotesize{$\pm$0.18} \end{tabular}
& \begin{tabular}{@{}c@{}}{55.83} \\ \footnotesize{$\pm$0.50} \end{tabular}
& \begin{tabular}{@{}c@{}}{96.95} \\ \footnotesize{$\pm$0.04} \end{tabular}
& \begin{tabular}{@{}c@{}}{81.99} \\ \footnotesize{$\pm$0.47} \end{tabular}
& \begin{tabular}{@{}c@{}}{63.39} \\ \footnotesize{$\pm$1.95} \end{tabular}
& \begin{tabular}{@{}c@{}}{68.68} \\ \footnotesize{$\pm$0.46} \end{tabular}
& \begin{tabular}{@{}c@{}}{82.67} \\ \footnotesize{$\pm$0.46} \end{tabular}
& \begin{tabular}{@{}c@{}}{68.19} \\ \footnotesize{$\pm$0.39} \end{tabular}
& \begin{tabular}{@{}c@{}}{72.48} \\ \footnotesize{$\pm$0.23} \end{tabular}
& \begin{tabular}{@{}c@{}}{53.17} \\ \footnotesize{$\pm$0.56} \end{tabular}
& \begin{tabular}{@{}c@{}}{70.75} \\ \footnotesize{$\pm$0.31} \end{tabular}
\\
Sub-word
& \begin{tabular}{@{}c@{}}{64.09} \\ \footnotesize{$\pm$0.28} \end{tabular}
& \begin{tabular}{@{}c@{}}{54.88} \\ \footnotesize{$\pm$0.39} \end{tabular}
& \begin{tabular}{@{}c@{}}{97.14} \\ \footnotesize{$\pm$0.03} \end{tabular}
& \begin{tabular}{@{}c@{}}{81.94} \\ \footnotesize{$\pm$0.28} \end{tabular}
& \begin{tabular}{@{}c@{}}{62.67} \\ \footnotesize{$\pm$2.87} \end{tabular}
& \begin{tabular}{@{}c@{}}{69.25} \\ \footnotesize{$\pm$0.42} \end{tabular}
& \begin{tabular}{@{}c@{}}{83.20} \\ \footnotesize{$\pm$0.27} \end{tabular}
& \begin{tabular}{@{}c@{}}{69.03} \\ \footnotesize{$\pm$0.44} \end{tabular}
& \begin{tabular}{@{}c@{}}{72.78} \\ \footnotesize{$\pm$0.13} \end{tabular}
& \begin{tabular}{@{}c@{}}{53.32} \\ \footnotesize{$\pm$0.44} \end{tabular}
& \begin{tabular}{@{}c@{}}{70.83} \\ \footnotesize{$\pm$0.35} \end{tabular}
\\
SubChar-Wubi
& \begin{tabular}{@{}c@{}}{63.89} \\ \footnotesize{$\pm$0.25} \end{tabular}
& \begin{tabular}{@{}c@{}}{58.64} \\ \footnotesize{$\pm$0.27} \end{tabular}
& \begin{tabular}{@{}c@{}}{97.02} \\ \footnotesize{$\pm$0.04} \end{tabular}
& \begin{tabular}{@{}c@{}}{81.70} \\ \footnotesize{$\pm$0.29} \end{tabular}
& \begin{tabular}{@{}c@{}}{64.61} \\ \footnotesize{$\pm$2.09} \end{tabular}
& \begin{tabular}{@{}c@{}}{68.75} \\ \footnotesize{$\pm$0.59} \end{tabular}
& \begin{tabular}{@{}c@{}}{82.81} \\ \footnotesize{$\pm$0.46} \end{tabular}
& \begin{tabular}{@{}c@{}}{68.93} \\ \footnotesize{$\pm$0.38} \end{tabular}
& \begin{tabular}{@{}c@{}}{72.54} \\ \footnotesize{$\pm$0.15} \end{tabular}
& \begin{tabular}{@{}c@{}}{54.68} \\ \footnotesize{$\pm$0.77} \end{tabular}
& \begin{tabular}{@{}c@{}}{71.36} \\ \footnotesize{$\pm$0.23} \end{tabular}
\\
SubChar-Pinyin
& \begin{tabular}{@{}c@{}}{63.68} \\ \footnotesize{$\pm$0.25} \end{tabular}
& \begin{tabular}{@{}c@{}}{58.81} \\ \footnotesize{$\pm$0.28} \end{tabular}
& \begin{tabular}{@{}c@{}}{97.04} \\ \footnotesize{$\pm$0.03} \end{tabular}
& \begin{tabular}{@{}c@{}}{81.74} \\ \footnotesize{$\pm$0.24} \end{tabular}
& \begin{tabular}{@{}c@{}}{65.90} \\ \footnotesize{$\pm$1.45} \end{tabular}
& \begin{tabular}{@{}c@{}}{68.89} \\ \footnotesize{$\pm$0.42} \end{tabular}
& \begin{tabular}{@{}c@{}}{82.87} \\ \footnotesize{$\pm$0.40} \end{tabular}
& \begin{tabular}{@{}c@{}}{67.98} \\ \footnotesize{$\pm$0.45} \end{tabular}
& \begin{tabular}{@{}c@{}}{73.06} \\ \footnotesize{$\pm$0.13} \end{tabular}
& \begin{tabular}{@{}c@{}}{53.03} \\ \footnotesize{$\pm$0.47} \end{tabular}
& \begin{tabular}{@{}c@{}}{71.42} \\ \footnotesize{$\pm$0.19} \end{tabular}
\\
    
    \hline 
    \multicolumn{12}{c}{\textit{12-layer, 2.3G Corpus}} \\
    \hline 
    CharTokenizer
    & \begin{tabular}{@{}c@{}}{64.39} \\ \footnotesize{$\pm$0.13} \end{tabular}
    & \begin{tabular}{@{}c@{}}{58.52} \\ \footnotesize{$\pm$0.46} \end{tabular}
    & \begin{tabular}{@{}c@{}}{97.02} \\ \footnotesize{$\pm$0.03} \end{tabular}
    & \begin{tabular}{@{}c@{}}{83.49} \\ \footnotesize{$\pm$0.38} \end{tabular}
    & \begin{tabular}{@{}c@{}}{68.09} \\ \footnotesize{$\pm$1.59} \end{tabular}
    & \begin{tabular}{@{}c@{}}{69.00} \\ \footnotesize{$\pm$0.35} \end{tabular}
    & \begin{tabular}{@{}c@{}}{82.77} \\ \footnotesize{$\pm$0.33} \end{tabular}
    & \begin{tabular}{@{}c@{}}{70.40} \\ \footnotesize{$\pm$0.34} \end{tabular}
    & \begin{tabular}{@{}c@{}}{74.44} \\ \footnotesize{$\pm$0.17} \end{tabular}
    & \begin{tabular}{@{}c@{}}{54.22} \\ \footnotesize{$\pm$0.40} \end{tabular}
    & \begin{tabular}{@{}c@{}}{72.23} \\ \footnotesize{$\pm$0.26} \end{tabular}
    \\
    SubChar-Pinyin
    & \begin{tabular}{@{}c@{}}{64.19} \\ \footnotesize{$\pm$0.14} \end{tabular}
    & \begin{tabular}{@{}c@{}}{59.67} \\ \footnotesize{$\pm$0.23} \end{tabular}
    & \begin{tabular}{@{}c@{}}{97.12} \\ \footnotesize{$\pm$0.03} \end{tabular}
    & \begin{tabular}{@{}c@{}}{82.28} \\ \footnotesize{$\pm$0.16} \end{tabular}
    & \begin{tabular}{@{}c@{}}{71.71} \\ \footnotesize{$\pm$2.03} \end{tabular}
    & \begin{tabular}{@{}c@{}}{69.30} \\ \footnotesize{$\pm$0.24} \end{tabular}
    & \begin{tabular}{@{}c@{}}{82.23} \\ \footnotesize{$\pm$0.27} \end{tabular}
    & \begin{tabular}{@{}c@{}}{70.43} \\ \footnotesize{$\pm$0.25} \end{tabular}
    & \begin{tabular}{@{}c@{}}{74.82} \\ \footnotesize{$\pm$0.09} \end{tabular}
    & \begin{tabular}{@{}c@{}}{55.92} \\ \footnotesize{$\pm$0.26} \end{tabular}
    & \begin{tabular}{@{}c@{}}{72.87} \\ \footnotesize{$\pm$0.17} \end{tabular}
    \\
    \hline 
    \multicolumn{12}{c}{\textit{12-layer, 22.1G Corpus}} \\
    \hline 
    CharTokenizer
    & \begin{tabular}{@{}c@{}}{64.43} \\ \footnotesize{$\pm$0.57} \end{tabular}
    & \begin{tabular}{@{}c@{}}{59.10} \\ \footnotesize{$\pm$0.29} \end{tabular}
    & \begin{tabular}{@{}c@{}}{97.12} \\ \footnotesize{$\pm$0.01} \end{tabular}
    & \begin{tabular}{@{}c@{}}{82.70} \\ \footnotesize{$\pm$0.02} \end{tabular}
    & \begin{tabular}{@{}c@{}}{70.39} \\ \footnotesize{$\pm$1.32} \end{tabular}
    & \begin{tabular}{@{}c@{}}{69.39} \\ \footnotesize{$\pm$0.06} \end{tabular}
    & \begin{tabular}{@{}c@{}}{82.97} \\ \footnotesize{$\pm$0.28} \end{tabular}
    & \begin{tabular}{@{}c@{}}{69.37} \\ \footnotesize{$\pm$0.14} \end{tabular}
    & \begin{tabular}{@{}c@{}}{76.34} \\ \footnotesize{$\pm$0.62} \end{tabular}
    & \begin{tabular}{@{}c@{}}{54.84} \\ \footnotesize{$\pm$1.24} \end{tabular}
    & \begin{tabular}{@{}c@{}}{72.81} \\ \footnotesize{$\pm$0.18} \end{tabular}
    \\
    SubChar-Pinyin
    & \begin{tabular}{@{}c@{}}{64.64} \\ \footnotesize{$\pm$0.47} \end{tabular}
    & \begin{tabular}{@{}c@{}}{59.14} \\ \footnotesize{$\pm$0.17} \end{tabular}
    & \begin{tabular}{@{}c@{}}{97.10} \\ \footnotesize{$\pm$0.04} \end{tabular}
    & \begin{tabular}{@{}c@{}}{83.56} \\ \footnotesize{$\pm$0.18} \end{tabular}
    & \begin{tabular}{@{}c@{}}{72.36} \\ \footnotesize{$\pm$0.98} \end{tabular}
    & \begin{tabular}{@{}c@{}}{70.67} \\ \footnotesize{$\pm$0.66} \end{tabular}
    & \begin{tabular}{@{}c@{}}{82.94} \\ \footnotesize{$\pm$0.05} \end{tabular}
    & \begin{tabular}{@{}c@{}}{69.50} \\ \footnotesize{$\pm$0.24} \end{tabular}
    & \begin{tabular}{@{}c@{}}{75.92} \\ \footnotesize{$\pm$0.45} \end{tabular}
    & \begin{tabular}{@{}c@{}}{58.64} \\ \footnotesize{$\pm$0.35} \end{tabular}
    & \begin{tabular}{@{}c@{}}{73.42} \\ \footnotesize{$\pm$0.09} \end{tabular}
    \\
    \bottomrule
    \end{tabular}
    }
    \end{adjustbox}
    \caption{Results on downstream datasets of different tokenizers. The last column indicates average performance. The subscript is the standard deviation. Models trained with sub-character tokenizers can match the performance of baseline models across all datasets. Ablation shows that increasing the model size or pretraining corpus size can slightly improve downstream task performance. These ablation results support our overall conclusion that models trained with SubChar tokenizers can closely match or slightly outperform the baselines.}
    \label{tab:main_results}
    \end{table*}

\textbf{BERT pretraining}. We follow the training procedure of BERT~\cite{BERT} except that the next sentence prediction objective is removed. The pretraining process consists of two stages. The first stage uses a maximum sequence length of $128$ with a batch size of $8$K for $8$K steps. The second stage uses a maximum sequence length of $512$ with a batch size of $4$K for $2$K steps. We experiment primarily with  $6$-layer Transformer~\cite{Transformer} models. To ablate the impact of model size, we also pretrain 12-layer Transformer models for the baseline CharTokenizer and proposed SubChar-Pinyin tokenizer. 
Other model configurations are the same for all models: $12$ attention heads, an intermediate size of $3072$, and a hidden size of $768$.

\textbf{BERT finetuning}.
For the finetuning on downstream datasets, 
we use a batch size of 32, maximum training epochs of 24 and tune max sequence length in \{96, 256, 512\}. Since the original test sets are not released, we use the original dev sets as the test sets and randomly hold-out 10\% of the training set as the dev sets. We select the best checkpoint on the dev sets and report performance on test sets.
These hyper-parameters are consistent with previous work.
For all experiments in this paper, we report the results of the average run of three different random seeds.
All experiments are done on NVIDIA A100 GPUs.

\section{Experiment Results}

In this section, we present the experiment results and the main findings. We not only evaluate on a wide range of common Chinese NLU datasets, but also perform robustness evaluation on both synthetic and real-world noisy data. 

\subsection{Standard Evaluation}
\label{sec:standard_eval}



We compare models trained with our SubChar tokenizers and the baseline tokenizers. There are multiple possible encoding methods for SubChar tokenizers as described in section~\ref{sec:method}. In this section, we choose two representative ones: \textit{Wubi} (glyph-based) and \textit{Pinyin} (pronunciation-based). We later show a full ablation of all different encoding methods in section~\ref{sec:encoding}.



Table~\ref{tab:main_results} shows the performance of BERT models with different tokenizers on downstream datasets. 
Examining the results of the 6-layer BERT models pretrained on the 2.3G Baidu Baike corpus, we observe that
despite some variation across different datasets, our proposed sub-character tokenizers can match the baselines on downstream datasets.
When scaling the 6-layer models to 12-layer, we observe moderate improvement on the average performance (70.75 $\rightarrow$ 72.23 for CharTokenizer and 71.42 $\rightarrow$ 72.87 for SubChar-Pinyin). 
Besides, we discuss the impact of pretraining data size in section~\ref{sec:pretraining-data}.
These results demonstrate that on standard NLU benchmarks, our proposed tokenizers can serve as a very strong alternative.

\begin{table*}[t]
    \centering
    \small
    \addtolength{\tabcolsep}{-2pt}
    \setlength{\tabcolsep}{3mm}{
    \begin{tabular}{ lcccccc }
    
    \toprule  
    \multicolumn{7}{c}{TNEWS} \\
    \hline
     & clean
     & 7.5 \%
     & 15.0 \%
     & 22.5 \% 
     & 30.0 \%
     & 37.5 \%
     \\
    \hline 
    CharTokenizer   & \textbf{64.10} & 63.09 & 58.96 & 50.91 & 38.33 & 25.20 \\
    Sub-word           & 64.09 & 62.82 & 57.75 & 48.67 & 36.37 & 25.72 \\
    SubChar-Pinyin          & 63.68 & 61.95 & 56.67 & 45.22 & 30.71 & 27.53 \\
    SubChar-Pinyin-NoIndex  & 63.28 & \textbf{63.28} & \textbf{63.28} & \textbf{63.28} & \textbf{63.28} & \textbf{63.28} \\ 
    \bottomrule
    \multicolumn{7}{c}{OCNLI} \\
    \hline
     & clean
     & 7.5 \%
     & 15.0 \%
     & 22.5 \% 
     & 30.0 \%
     & 37.5 \%
     \\
    \hline 
    CharTokenizer   & 68.37 & 64.89 & 56.85 & 47.65 & 40.48 & 36.36 \\
    Sub-word           & \textbf{68.84} & 64.33 & 56.49 & 48.07 & 42.68 & 38.28 \\
    SubChar-Pinyin          & 67.70 & 61.93 & 54.39 & 46.01 & 40.24 & 37.33 \\
    SubChar-Pinyin-NoIndex  & 67.91 & \textbf{67.91} & \textbf{67.91} & \textbf{67.91} & \textbf{67.91} & \textbf{67.91} \\
    
    \bottomrule
    \multicolumn{7}{c}{C3} \\
    \hline
     & clean
     & 7.5 \%
     & 15.0 \%
     & 22.5 \% 
     & 30.0 \%
     & 37.5 \%
     \\
    \hline 
    CharTokenizer   & 53.13 & 51.46 & 49.22 & 47.71 & 46.78 & 43.95 \\
    Sub-word           & 53.55 & 51.66 & 49.49 & 47.81 & 46.24 & 43.58 \\
    SubChar-Pinyin          & 52.87 & 50.45 & 47.26 & 44.50 & 42.42 & 40.07 \\
    SubChar-Pinyin-NoIndex  & \textbf{53.65} & \textbf{53.65} & \textbf{53.65} & \textbf{53.65} & \textbf{53.65} & \textbf{53.65} \\
    \bottomrule
    \end{tabular}}
    \caption{Results for noisy evaluation with homophone typos. Different columns correspond to different percentages of typos in the test data. The BERT model with our SubChar-Pinyin-NoIndex tokenizer (results in \textbf{bold}) suffers no performance drop on noisy test data since it is robust to all homophone typos.}
    \label{tab:noisy_results_phonology}
    \end{table*}

\begin{figure}[t]
\centering
\includegraphics[width=0.49\textwidth]{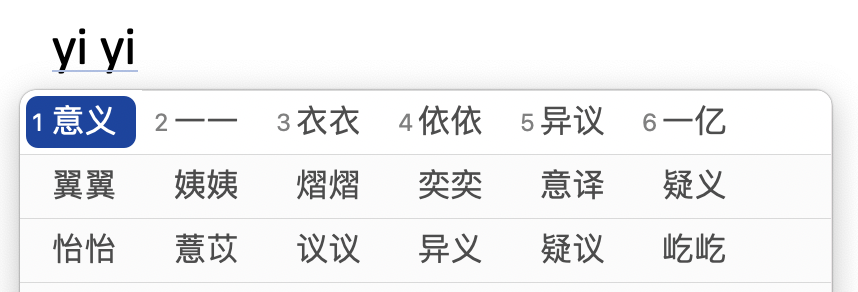}
\caption{An actual interface of the popular \textit{pinyin} input method. The first line \textit{\underline{yi yi}} is the user input of the romanization sequence, all words with this same pronunciation are listed below for users to choose from.}
\label{fig:pinyin}
\end{figure}

\begin{figure}[t]
\centering
\includegraphics[width=0.49\textwidth]{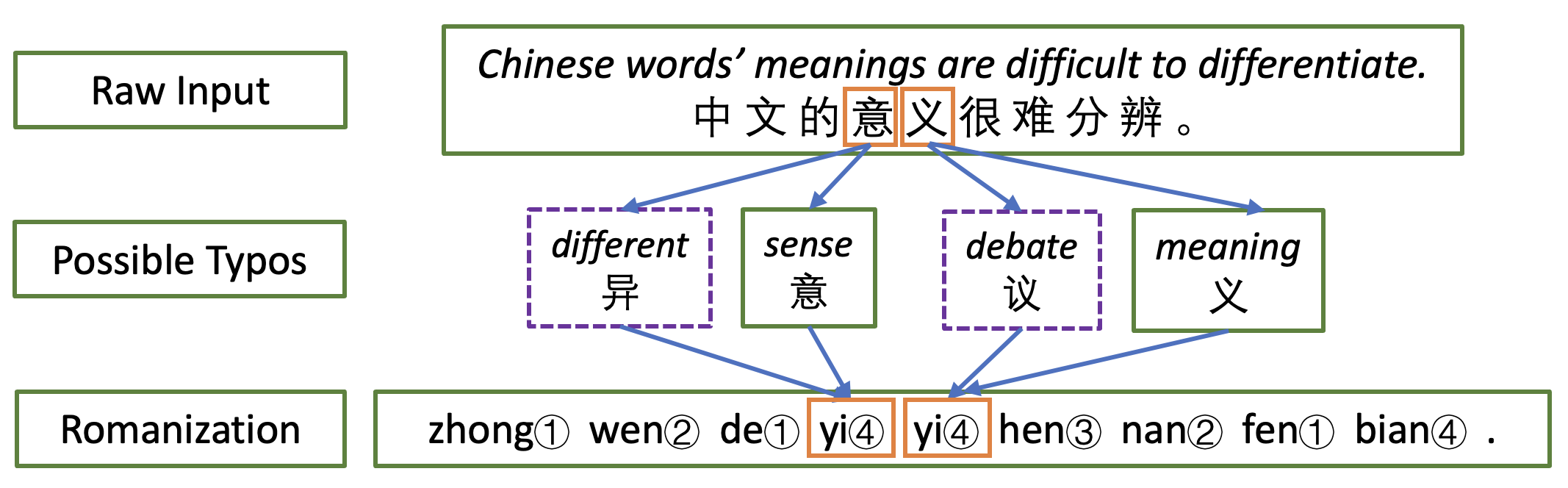}
\caption{Illustration of how our SubChar-Pinyin-NoIndex tokenizer is robust to any homophone typos. The possible homophone typos (characters in purple dashed boxes) are mapped into the same romanization sequence as the intended correct characters, and hence the resultant tokenization based on the romanized sequences would be the same.}
\label{fig:no_index}
\end{figure}

\subsection{Robustness Evaluation}
\label{sec:noisy_eval}

Apart from evaluating on the standard benchmarks, we also verify whether our proposed tokenization methods are better at handling noisy inputs.
We cover two major Chinese input methods: keyboard input and speech input. 
For keyboard input, we construct synthetic noise tests via character substitutions. 
For speech input, we use a noisy test set including inputs with diverse accents, which poses greater typo diversity.  
Our SubChar-Pinyin method shows advantage in both cases.

\paragraph{Synthetic Typos}
We simulate the homophone typos that are common in real-world Chinese writing systems, especially user-generated inputs. As shown in Figure~\ref{fig:pinyin}, \textit{pinyin} input is the most widely used keyboard input method for Chinese users.\footnote{\url{https://en.wikipedia.org/wiki/Chinese_input_methods_for_computers}} When users type in the romanization of the intended characters, the input interface will present all Chinese characters with the same romanization for the users to choose from. As a result, it is common for users to choose the wrong characters either by mistake or because they are unclear about the differences among these homophones.

In such cases, our SubChar-Pinyin-NoIndex tokenizer (described in section~\ref{sec:char_encode}) has the advantage of being robust towards any such homophone typos. As illustrated in Figure~\ref{fig:no_index}, the character encoding will map all homophones of a character into the same romanization sequence before undergoing the sub-word tokenization. As a result, the tokenized output will be identical no matter what the typo character is as long as it is a homophone of the intended character.


 We inject synthetic noises into the test data and examine whether models trained on clean training data can perform well on these noisy data.
To construct the noisy data, we replace the original correct characters with their homophones, \textit{e.g.}, change `意'\ (sense) to `异'\ (different)' and `义'\ (meaning) to `议'\ (debate).\footnote{Interestingly, all these four characters have the same pronunciation but different meanings. Moreover, ``意义"\ (meaning) and ``异议"\ (objection) are homophone words.}
Specifically, we randomly sample a certain ratio $r\%$ of the original characters. For each of them, we replace it with a randomly sampled homophone from all its homophones obtained via a Pinyin dictionary (no replacement if it has no homophones).




The results are shown in Table~\ref{tab:noisy_results_phonology}. 
We observe that there can be a significant drop in performance where there exist homophone typos in the test data. For example, the BERT model trained with CharTokenizer drops from $64.10\%$ accuracy on clean data to $25.20\%$ accuracy when $37.5\%$ of the characters in test inputs are replaced with homophone typos.
Overall, we find that the character tokenizer, sub-word tokenizer, as well as the vanilla SubChar-Pinyin tokenizer, cannot handle such noisy data. However, our SubChar-Pinyin-NoIndex tokenizer exhibits \textbf{no performance drop} under noises.
Moreover, despite learning a shared representation for homophones, the model with SubChar-Pinyin-NoIndex still performs competitively on the clean test sets, either match (on C3) or only a little worse than the baselines (on TNEWS and OCNLI).

\paragraph{Real-World Typos}
While the above synthetic typos aim to simulate typos in keyboard inputs, another major input method is through speech input where users speak to their devices (like mobile phones) and their speech input is then converted to text for downstream tasks. 
In order to evaluate model robustness in such scenarios, we use a realistically collected test set that captures such speech input typos. 
Specifically, we use the speech-noise version of the AFQMC test set from the READIN~\cite{READIN} benchmark. 
For each example in this noisy AFQMC test set, three annotators with different accents read the original input, and then the speech recordings are converted to text using commercial automatic speech recognition (ASR) software. We refer readers to the dataset description paper for more data construction details. 
When computing performance for each test example, we compute both the average across different annotations (Noisy-Average), as well as the worst performance across different annotations (Noisy-Worst), and then take the macro-average across all examples. The character-level error rate of the noisy test set is 30\% on average. 

 %

This AFQMC noisy test set contains not only homophone typos, but also a wide range of other types of real-world input noises due to both the accent variations and ASR errors. The greater diversity of typo types in the real-world test set makes it much more challenging to maintain robustness than the synthetic setting which only considers homophone typos. 
While the original AFQMC is a binary classification task that classifies whether the question pair is a paraphrase or not, we find that models trained on the AFQMC training set exploit spurious correlations like lexical overlap, even though we explicitly balanced the training set. In particular, when introducing typos in the test data, performance on positive examples drops drastically due to lower lexical overlap, while the performance on negative examples stays or even improves a little because of the lower lexical overlap caused by the typos. This is similar to previous findings on HANS~\cite{McCoy2019RightFT} and PAWS~\cite{Zhang2019PAWSPA}. Hence, we follow the evaluation practice when dealing with spurious correlation, which is to focus on improving the worst-group performance, and in this case, we focus on improving performance on the positive examples against the impact of typos.

\begin{table}[t]
    \addtolength{\tabcolsep}{-4pt}
    \centering
    \small
    \setlength{\tabcolsep}{1mm}{
    \begin{tabular}{ lccc }
    \toprule        & Clean  & N-Avg & N-Worst \\
    \hline 
    CharTokenizer   &  73.02 & 44.11 & 18.81 \\
    Sub-word & 74.22 & 42.21 & 16.91 \\
    SubChar-Pinyin & 73.24 & \textbf{45.24} & \textbf{19.47} \\
    \bottomrule
    \end{tabular}
    }
    \caption{Results on the real-world AFQMC noisy test set. Each clean test instance is annotated by three different annotators, we report both the macro-average on these noisy annotations (N-Average) as well as the average of the worst-case performance across all test examples (N-Worst). SubChar-Pinyin outperforms baselines on the challenging noisy test set (best results on the noisy test set are in bold).}
    \label{tab:afqmc_noisy}
    \vspace{-1em}
    \end{table}
The results are shown in Table~\ref{tab:afqmc_noisy} where we report performance on the AFQMC positive examples. All models are trained on the original clean data from AFQMC (we balanced the positive and negative classes during training). We evaluate on the original clean test set, the Noisy-Average performance (N-Average), and the Noisy-Worst performance (N-Worst). We can see that despite this more challenging speech typo setting, our SubChar-Pinyin model still outperforms the baselines. 

\begin{table*}[t]
    \centering
    \small
    \begin{adjustbox}{width=\linewidth,center}
    \addtolength{\tabcolsep}{-2pt}
    \setlength{\tabcolsep}{2mm}{
    \begin{tabular}{ lcccccccl }
    \toprule 
     & TNEWS
     & IFLYTEK
     & CLUEWSC
     & AFQMC
     & CSL
     & OCNLI
     & C3
     & AVG
     \\
     \hline  
    Sub-word                & 64.09 & 54.88 & 62.67 & 69.25 & 83.20 & 69.03 & 53.32 & 65.21 \\
    Sub-word + \cws{}       & 64.26 & 54.15 & 63.05 & 69.62 & 82.87 & 68.64 & 51.77 & 64.91 (-0.30) \\
    SubChar-Wubi            & 63.89 & 58.64 & 64.61 & 68.75 & 82.81 & 68.93 & 54.68 & 66.04 \\
    SubChar-Wubi + \cws{}   & 63.57 & 58.01 & 64.38 & 69.41 & 82.62 & 69.43 & 53.15 & 65.80 (-0.24) \\
    SubChar-Pinyin          & 63.68 & 58.81 & 65.90 & 68.89 & 82.87 & 67.98 & 53.03 & 65.88 \\
    SubChar-Pinyin + \cws{} & 63.73 & 57.89 & 64.51 & 69.66 & 82.90 & 69.93 & 53.63 & 66.04 (+0.16) \\
    \bottomrule
    \end{tabular}}
    \end{adjustbox}
    \caption{Results of models trained with different tokenizers. Numbers in brackets indicate the difference between adding and not adding the \cws{} step in tokenization. Adding \cws{} brings \textbf{no} significant improvement in performance.}
    \label{tab:cws_results}
    \end{table*}

These results highlight the robustness advantage of our Sub-Character tokenization method in both dealing with synthetic homophone typos as well as on more diverse real-world typos.

\begin{table}[t]
    \addtolength{\tabcolsep}{-4pt}

    \centering
    \small
    \setlength{\tabcolsep}{1mm}{
    \begin{tabular}{ lcc }
    \toprule        & CMRC  & CLUENER \\
    \hline 
    CharTokenizer   & 56.58 & 69.61  \\
    Sub-word   & 55.85 & 67.94  \\
    SubChar-Wubi    & 54.45 & 70.63  \\
    SubChar-Pinyin  & 55.18 & 70.77 \\
    \bottomrule
    \end{tabular}
    }
    \caption{Results on two character-level classification datasets: CMRC (span-extraction) and CLUENER (named entity recognition). Models are 6-layer BERT. Models with SubChar tokenizers perform close to the baseline models.}
    \label{tab:cmrc_ner}
    \vspace{-1em}
    \end{table}

\subsection{Effect of CWS}

 We examine the impact of incorporating \cws{} in the tokenization as described in Section~\ref{sec:cws}.
We train tokenizers with and without \cws{} and compare the performance of the corresponding pretrained models. As shown in Table~\ref{tab:cws_results}, we can see that adding \cws{} as an additional step does not help downstream task performance. These results serve as empirical evidence that \cws{} is ineffective in the use of \plm{}s, complementary to the results of \citet{CWSnecessary} on models without pretraining.

\subsection{Character-Level Tasks}

The evaluation in Section~\ref{sec:standard_eval} is restricted to sequence-level classification tasks such as single-sentence classification, sentence-pair classification and machine reading comprehension. 

One might wonder how do SubChar tokenizers handle character-level tasks where classification is done on every single character, such as sequence labeling and span extraction. Since SubChar tokenizers may combine multiple characters into one token or split one character into sub-character tokens, directly adding a classification head on each token may cause discrepancy with the human annotation, which is done on the character level. For example, it is infeasible to evaluate the POS tag of a sub-character token.




To handle such situations, we perform classification on the character level for these tasks.
To obtain the representation of each character, we average the representations of all its sub-character tokens. We apply this on the final layer of BERT and feed the character representation to a linear classifier for downstream tasks.


We measure the performance of this approach on CMRC (span-extraction reading comprehension) and CLUENER (named entity recognition) and show the results in Table~\ref{tab:cmrc_ner}. The results show that our model can indeed handle character-level tasks with this simple adaptation. There might be better ways of adopting our model on character-level tasks, and we leave it to future work.


\section{Analysis}

\label{sec:analysis}

\begin{figure}[t]
\centering
\includegraphics[width=0.47\textwidth]{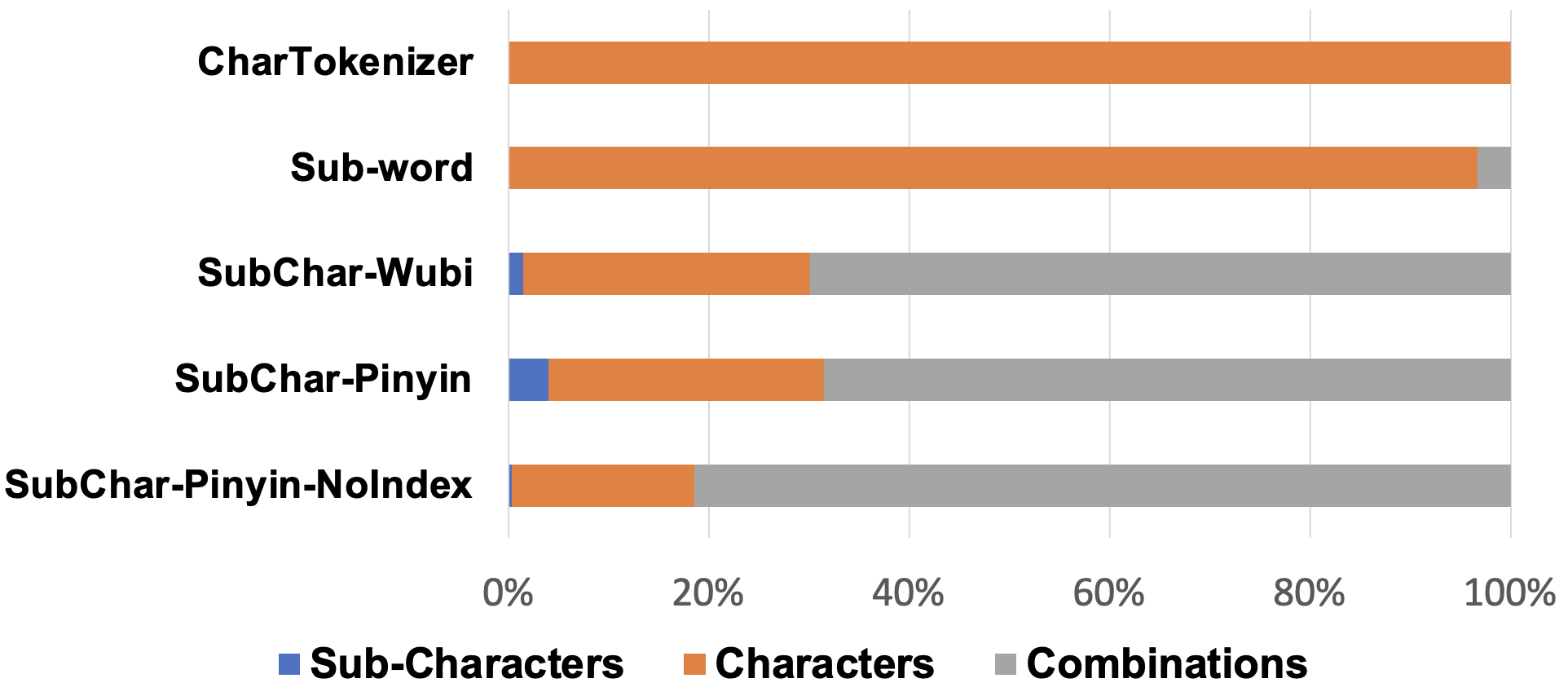}
\caption{Breakdown of different types of tokens in the vocabularies of various tokenizers. We observe the clear trend that in our SubChar tokenizers, a small fraction of sub-character tokens saves the space to store much more character combination tokens (e.g., words and phrases).}
\label{fig:composition}
\end{figure}

In this section, we conduct various analyses to better understand the working mechanisms of SubChar tokenization, including illustrations of the efficiency improvement and ablations on different components of our tokenization pipeline. 

\subsection{Vocabulary Composition}

We break down the vocabulary of each tokenizer into three different categories: sub-character tokens, character tokens, and character combination tokens (words and phrases). 
As shown in Figure~\ref{fig:composition}, character tokenizers only have character tokens, while sub-word tokenizers have a small percentage of combination tokens. The main reason for the relatively small number of combination tokens in sub-word tokenizers is that unlike how English words are composed with $26$ alphabet letters, there are thousands of unique Chinese characters, which take up a large proportion of the vocabulary in order to maintain coverage.

In contrast, SubChar tokenizers use a very small fraction of sub-character tokens to compose many complex Chinese characters, therefore saving up a large percentage of the vocabulary to store combination tokens. This brings the advantage of having more words and phrases in the tokenized outputs, thus shortening the sequence lengths, as elaborated in the next section.

\subsection{Efficiency Improvement}

\begin{table}[t]
    \centering
    \small
    \addtolength{\tabcolsep}{-4pt}
    \setlength{\tabcolsep}{1mm}{
    \begin{tabular}{ lcc }
    \toprule 
     & iFLYTEK
     & TNEWS
     \\
\hline
CharTokenizer & 289.0 & 22.0 \\
Sub-word & 255.2 & 20.1 \\
SubChar-Wubi & 183.2  & 15.8 \\
SubChar-Pinyin & 185.2  &  16.1 \\
SubChar-Pinyin-NoIndex & \textbf{175.4} & \textbf{15.2} \\
    \bottomrule
    \end{tabular}}
    \caption{Comparison of average length of tokenized sequences with different tokenizers. 
    SubChar tokenizers produce much shorter tokenized sequences than the baselines. SubChar-Pinyin-NoIndex tokenizer achieves the most length reduction. BPE and Unigram LM counterparts achieve similar speedup improvement. }
    \label{tab:output_lengths}
    \end{table}

\begin{table}[t]
    \centering
    \small
    \addtolength{\tabcolsep}{-3pt}
    \setlength{\tabcolsep}{1mm}{
    \begin{tabular}{ lcc }
    \toprule 
                        & TNEWS     & iFLYTEK \\
    \hline
    CharTokenizer       & 100.0\%     & 100.0\% \\
    Sub-word            & 99.9\%    & 92.6\% \\
    SubChar-Wubi        & 87.0\%    & 69.6\% \\
    SubChar-Pinyin      & 83.8\%    & 70.4\% \\
    SubChar-Pinyin-NoIndex & \textbf{82.7\%} & \textbf{68.9\%} \\

    \bottomrule
    \end{tabular}}
    \caption{Finetuning time of models with different tokenizers. Numbers indicate time relative to the CharTokenizer baseline model. Models with SubChar tokenizers take much shorter time to finish finetuning. SubChar-Pinyin-NoIndex brings the most speedup. }
    \label{tab:finetune_time}
\end{table}

The direct consequence of having more character combinations in the vocabulary is that the tokenized sequences are shorter. Table~\ref{tab:output_lengths} shows the average sequence length by using different tokenizers on two downstream datasets. 
We observe that SubChar tokenizers can tokenize the inputs into much shorter sequences.

Moreover, our SubChar tokenizers can speed up training for both pretraining and finetuning.
During finetuning, we can pack multiple sequences into one input sequence to reduce the computation waste introduced by sequence padding~\cite{kosec2021packing}, and shorter sequence lengths allow the sequences to be packed more densely, thus increasing the overall throughput.

 \begin{figure}[t]
\centering
\includegraphics[width=0.45\textwidth]{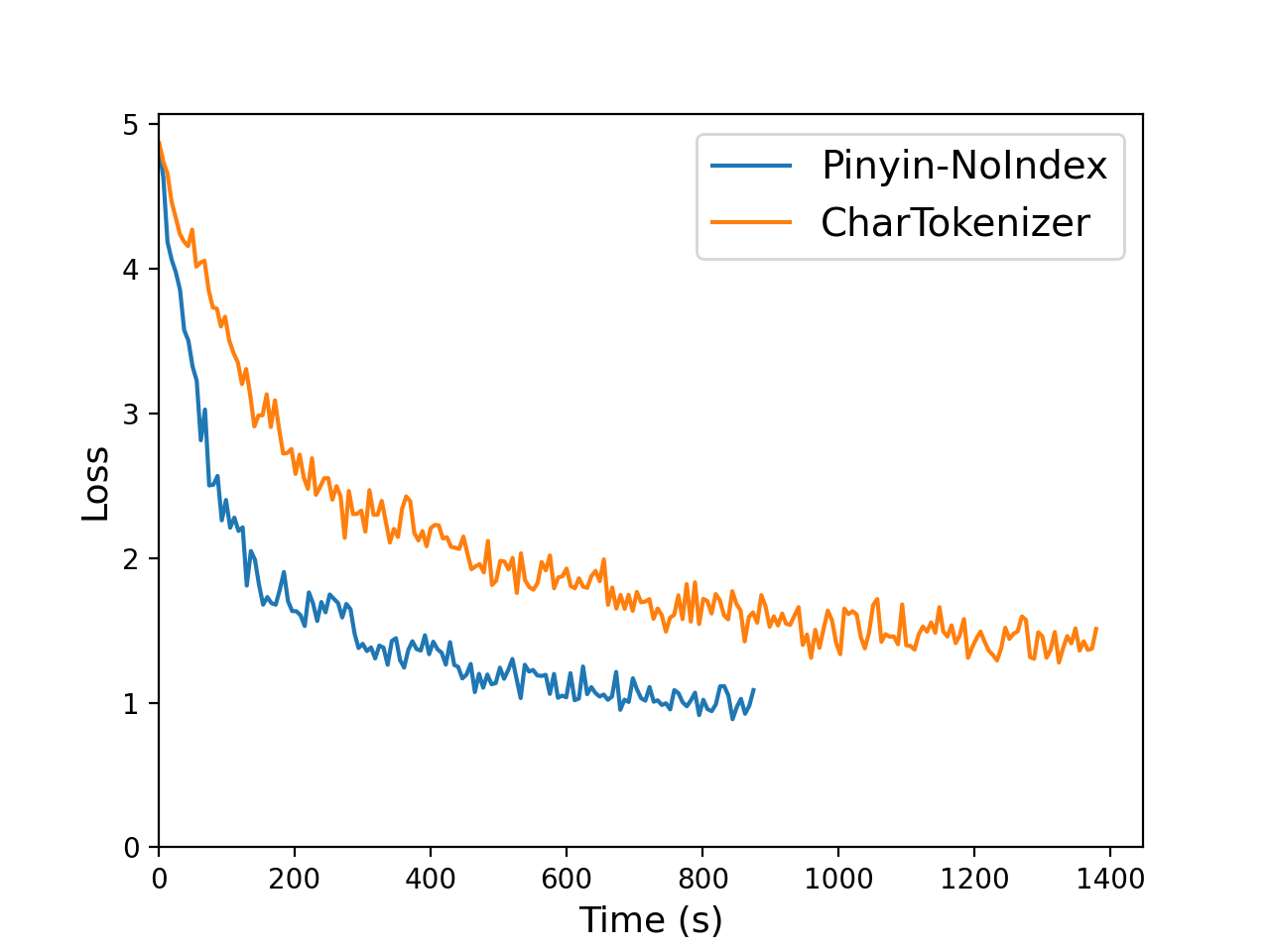}
\caption{Training curves on the iFLYTEK dataset with two different models. The y-axis indicates classification loss (cross-entropy), the x-axis indicates time (seconds). Our SubChar-Pinyin-NoIndex model gets a lower loss than the CharTokenizer baseline throughout training.}
\label{fig:ifly_loss_time}
\end{figure}

\begin{table}[t]
    \centering
    \small
    \addtolength{\tabcolsep}{-2pt}
    \setlength{\tabcolsep}{2mm}{
    \begin{tabular}{ lc }
    \toprule 
      & Tokenized Corpus Size 
     \\
\hline
CharTokenizer & 100.0\% \\
Sub-word & 91.4\%  \\
SubChar-Wubi & 77.2\% \\
SubChar-Pinyin & 78.4\% \\
SubChar-Pinyin-NoIndex & \textbf{74.7\%} \\

    \bottomrule
    \end{tabular}}
    \caption{Relative size (disk memory) of the tokenized pretraining corpus with different tokenizers. SubChar tokenizers produce much smaller tokenized corpus due to their ability to tokenize inputs into shorter sequences.}
    \label{tab:compression}
    \end{table}

Table~\ref{tab:finetune_time} shows the model finetuning time relative to the CharTokenizer baseline. We observe significant speedup by SubChar tokenizers, finishing in as little as $68.9\%$ time on iFLYTEK with the SubChar-Pinyin-NoIndex tokenizer.
In Figure~\ref{fig:ifly_loss_time}, we plot the training curves for the CharTokenizer baseline and the SubChar-Pinyin-NoIndex model on the iFLYTEK dataset, we observe that our SubChar-Pinyin-Noindex model indeed converges much faster and achieves lower training loss in the end.

The speedup on pretraining is also significant. While the running speed differs on different machines, the compression brought by the shorter tokenized outputs is hardware-invariant. In Table~\ref{tab:compression}, we show the relative size (disk memory) of the tokenized pretraining corpus. We observe that SubChar tokenizers can tokenize the raw pretraining texts into shorter sequences than the baselines, thus resulting in a much smaller pretraining data (\textit{e.g.}, as much as $25.3\%$ smaller than that of the CharTokenizer baseline with SubChar-Pinyin-NoIndex). In turn, this can translate to much faster pretraining on any training infrastructure.


\begin{table}[t]
    \centering
    \small
    \addtolength{\tabcolsep}{-4pt}
    \setlength{\tabcolsep}{1mm}{
    \begin{tabular}{ lcc }
    \toprule 
     & iFLYTEK
     & TNEWS
     \\
\hline
\multicolumn{3}{l}{\textit{Vocab Size = 22675}} \\
Sub-word & 255.2 & 20.1 \\
SubChar-Pinyin-NoIndex & \textbf{175.4} & \textbf{15.2} \\
\midrule
\multicolumn{3}{l}{\textit{Vocab Size = 40000}} \\
Sub-word & 188.9 & 15.9 \\
SubChar-Pinyin-NoIndex & \textbf{166.1} & \textbf{14.4} \\
\midrule
\multicolumn{3}{l}{\textit{Vocab Size = 60000}} \\
Sub-word & 176.2 & 14.9 \\
SubChar-Pinyin-NoIndex & \textbf{164.0} & \textbf{14.1} \\
    \bottomrule
    \end{tabular}}
    \caption{Comparison of average length of tokenized sequences with different tokenizers and different vocabulary sizes.}
    \label{tab:vocab_size_ablation}
    \end{table}

\begin{table*}[t]
    \centering
    \small
    \addtolength{\tabcolsep}{-2pt}
    \setlength{\tabcolsep}{3.5pt}{
    \begin{tabular}{ lcccccccc }
    \toprule             & TNEWS & IFLY  & BQ    & WSC   & AFQMC & CSL   & OCNLI & AVG \\
    \hline
    SubChar-Pinyin       & 63.68 & 58.81 & 81.74 & 65.90 & 68.89 & 82.87 & 67.98 & 70.16 \\
    SubChar-Zhuyin       & 64.91 & 59.39 & 81.41 & 62.72 & 69.14 & 82.60 & 69.12 & 69.90 \\
    SubChar-Stroke       & 64.26 & 55.44 & 81.52 & 62.06 & 69.88 & 83.16 & 68.98 & 69.33 \\
    SubChar-Wubi         & 63.81 & 58.74 & 81.55 & 64.61 & 69.66 & 82.44 & 68.02 & 69.90 \\
    SubChar-Zhengma      & 63.86 & 59.51 & 81.59 & 63.27 & 70.47 & 82.91 & 69.03 & 70.09 \\
    SubChar-Cangjie      & 64.10 & 57.77 & 81.98 & 62.39 & 68.95 & 82.60 & 68.46 & 69.46 \\
    SubChar-Byte         & 63.58 & 59.55 & 81.65 & 63.60 & 68.60 & 82.66 & 67.93 & 69.65 \\
    SubChar-RandomIndex  & 64.11 & 59.16 & 81.64 & 63.93 & 68.53 & 82.86 & 69.39 & 69.95 \\
    \midrule
    SubChar-Pinyin (BPE) & 63.86 & 58.84 & 82.12 & 65.57 & 69.86 & 82.86 & 68.57 & 70.24 \\
    \bottomrule
    \end{tabular}}
    \caption{Results of SubChar tokenizers when using different encoding methods. The last row is a model with SubChar-Pinyin tokenizer using BPE as the subword tokenization algorithm, all previous rows are using unigram LM as the subword tokenization implementation. All models have 6-layers with the same hyper-parameters. The impact of different encoding methods on downstream performance is small, and the ULM and BPE versions of SubChar-Pinyin also achieve similar results.}
    \label{tab:encode_ablation}
    \end{table*}

\subsection{Impact of Vocabulary Size}

Intuitively, when we increase the vocabulary size, there will also be more room to store combination tokens (\textit{e.g.}, words and phrases), leading to a decrease in tokenization length and thus better efficiency. Although we used the standard vocabulary size of 22675 in our previous experiments, to understand whether the efficiency benefits of SubChar tokenization wear off at larger vocabulary size, we perform an additional ablation on the impact of vocabulary size. 

As shown in Table~\ref{tab:vocab_size_ablation}, as we increase the vocabulary size, the efficiency advantage of SubChar tokenizers slightly diminishes. However, even at a very large vocab size of 60000, our SubChar-Pinyin tokenizer still tokenizes the inputs into significantly shorter sequences than the Sub-word baseline.  We thus conclude that the efficiency advantage of our SubChar tokenizers would hold in most practical cases where the vocabulary size is typically under 60000 (such as BERT and RoBERTa).

\subsection{Impact of Pretraining Data Size}
\label{sec:pretraining-data}

To understand the impact of pretraining data size, we take the checkpoints of the 12-layer Transformer models pretrained on the 2.3G Baike corpus, and further pretrain them on a much larger corpus of 22.1GB text. This 22.1GB corpus is sampled from Chinese web text\footnote{\url{https://github.com/OpenBMB/CPM-Live}}, mainly consisting of books and web pages. We further pretrain for 8K steps with a maximum sequence length of 512. 

As shown in the bottom block of Table~\ref{tab:main_results}, further training on this larger corpus leads to small improvement on average performance (72.23 $\rightarrow$ 72.81 for CharTokenizer and 72.87 $\rightarrow$ 73.42 for SubChar-Pinyin), possibly because the original models trained on 2.3GB corpus are already close to being fully trained. More importantly, this result shows that even with pretraining on larger corpora, our proposed methods can still match or slightly outperform baselines on the downstream datasets. 

\subsection{Impact of Encoding Methods}
\label{sec:encoding}

As described in Section~\ref{sec:method}, 
we experiment with different types of encoding methods and compare their downstream performance to analyze the impact.

Our previous encoding methods are based on the hypothesis that linguistic information such as glyph or pronunciation provides useful inductive biases to the model. However, in the case where this hypothesis is not true, it is possible that non-linguistic encoding methods may work as well. To verify this, we add two encoding methods that do not consider any linguistic information: \textit{Byte Encoding} and \textit{Random Index Encoding}, for the purpose of ablation analysis.

In \textit{Byte Encoding}, we convert every character into its byte sequence, same as in ByT5~\cite{ByT5}. In cases where the byte sequence consists of multiple indices (each Chinese character has three byte indices), we concatenate them and append the character separation symbol as the encoding (\textit{e.g., 魑} $\rightarrow$ \textit{233\_173\_145\#}). 

In \textit{Random Index Encoding}, we map each character into a unique and randomly generated five-digit index and append the character separation symbol as the encoding ( \textit{e.g., 魑} $\rightarrow$ \textit{29146\#} ).

We train SubChar tokenizers with all the different encoding methods and compare the corresponding BERT models using these tokenizers on downstream tasks. The results are presented in Table~\ref{tab:encode_ablation}. We observe that the differences between these different tokenizers are rather small in terms of the model performance on downstream datasets. Moreover, perhaps somewhat surprisingly, tokenizers with the non-linguistic encoding methods -- SubChar-Byte and SubChar-RandomIndex -- can also perform competitively despite the fact that they do not capture glyph or pronunciation information like the other tokenizers. 

These results suggest that linguistic encoding may not be necessary for SubChar tokenizers to achieve high performance on downstream tasks. However, the linguistic encoding methods can build more robust and efficient tokenizers as illustrated in previous sections.





\subsection{Impact of Vocabulary Construction Algorithm}
\label{sec:bpe}

In previous experiments, we used the Unigram LM implementation in SentencePiece for vocabulary construction. We perform an additional ablation where we replace Unigram LM with Byte Pair Encoding (BPE) for vocabulary construction to train a pinyin-based tokenizer, while holding all other hyper-parameters constant. 

We compare the SubChar-Pinyin-BPE variant with the unigram LM (SubChar-Pinyin) tokenizer. We find that these two perform similarly. In terms of \textbf{efficiency}: SubChar-Pinyin-BPE tokenizes iFLYTEK to an average length of 184.4 and tokenizes TNEWS to an average length of 15.9. In comparison, SubChar-Pinyin tokenizes iFLYTEK to an average length of 185.2 and tokenizes TNEWS to an average length of 16.1. The vocabulary compositions of the two are also similar, where character combination takes up the majority of the space in the vocabulary for both BPE and unigram LM implementations. 
In terms of \textbf{performance}, we observe in Table~\ref{tab:encode_ablation} that the BPE implementation and the unigram LM implementation have little difference in downstream task performance.
Based on these results, we conclude that the choice of which vocabulary construction to use has a marginal impact on the tokenization efficiency and model performance. 


\section{Related Work}
\paragraph{Chinese \plm{}s.} Chinese BERT~\cite{BERT} is the first Chinese \plm, which adopts the character tokenization. Then, researchers have explored techniques to explicitly incorporate the word-level information into Chinese \plm{}s for better performance. \citet{MVP-BERT} and \citet{AMBERT} expand BERT vocabulary with Chinese words apart from Chinese characters and incorporate them in the pretraining objectives. \citet{WWM}, \citet{NEZHA}, and \citet{ERNIE-GRAM} consider coarse-grained information through masking whole words and $n$-grams during the masked language modeling pretraining. \citet{ZEN} incorporate word-level information via superimposing the character and word embeddings. \citet{Lattice-BERT} incorporate Chinese word lattice structures in pretraining. Different from these studies, we investigate the information in the sub-character level for Chinese \plm{}s.
  
\paragraph{Linguistically-Informed Techniques for Chinese NLP.} 
Before the era of \plm{}, many efforts have been made to incorporate linguistic knowledge, including both glyph~\cite{sun2014radical,yu2017joint,cw2vec} and pronunciation~\cite{zhang2019learning,chaudhary-etal-2018-adapting}, into word embedding~\cite{mikolov2013distributed}.
Beyond word-level representation, researchers explore the use of linguistic information to enhance sequential models~\cite{dong2016character,bharadwaj-etal-2016-phonologically,liu-etal-2017-learning}, especially BERT~\cite{glyce,sun-etal-2021-chinesebert}.
Compared to these works, we do not incorporate additional information from sources like images, instead, our proposed tokenization methods are drop-in replacements to existing tokenizers, without adding any extra layers or parameters. 
Besides, \cws{} is a common preprocessing step for Chinese NLP~\cite{THULAC}, 
\citet{CWSnecessary} empirically analyze whether \cws{} is helpful for Chinese NLP tasks before the era of \plm{}s and find that the answer is no in many cases. 
In our work, we also spend a section examining the impact of \cws{} specifically for \plm{}s. 
Moreover, as shown by~\citet{huang-etal-2021-phmospell}, incorporating linguistic information also benefits spelling check.
Instead of explicitly using spelling check, our linguistically-informed tokenizations are robust to spelling errors.

\paragraph{Granularity of Tokenization.} Although sub-words are taken to be the default granularity of tokenization since the release of BERT, researchers also explore different granularities for \plm{}s. For instance, ELMo~\cite{ELMo}, the early pioneer of \plm{}s, starts by using character representation. \citet{CharBERT} combine character representations with sub-word representations for better performance and robustness. 
\citet{Nzeyimana2022KinyaBERTAM} incorporate a morphological analyzer for tokenization and achieve gains for the Kinyarwanda language model. 
More recently, there is a trend in tokenization-free methods, including Byte-BPE~\cite{BBPE}, CANINE~\cite{CANINE}, ByT5~\cite{ByT5}, and Charformer~\cite{Charformer}, which discard explicit tokenization and directly represent inputs as small units such as bytes. The downside of these tokenization-free approaches is obvious: the longer tokenized sequence lengths slow down both training and inference. Contrary to them, our sub-character tokenization encourages the use of more character combinations, which largely shortens the tokenized sequences.

\section{Conclusion}

In this work, we propose sub-character tokenization and conduct comprehensive experiments to illustrate its advantage over existing tokenization methods. 
%
%
Compared to treating each individual character as a token (CharTokenizer) or directly running sub-word tokenization on the raw Chinese text (sub-word tokenizer), our SubChar tokenizers not only perform competitively on downstream NLU tasks, more importantly, they can be much more efficient and robust.
We conduct a series of ablation and analysis to understand the reasons why SubChar tokenizers are more efficient, as well as the impact of linguistic and non-linguistic encoding.
Given the advantages of our SubChar tokenizers, we believe that they are better alternatives to all existing Chinese tokenizers, especially in applications where efficiency and robustness are critical. 
%
%
It is possible that our approach can be useful for other morphologically poor languages and more complicated methods could be developed based on SubChar tokenization for even better performance. We leave these interesting directions for future exploration.
On a broader level, our work makes an important attempt in developing more tailored methods for a language drastically different from English with promising results. We believe that this is a crucial future direction for the community given the language diversity in the world.
We hope that our work can inspire more such work in order to benefit language technology users from different countries and cultures.


\section*{Limitations}

Our experiments are focused on natural language understanding tasks. We recognize that adapting our SubChar tokenization to language generation tasks might require additional efforts, for example, we may want to avoid cases of predicting sub-character tokens that do not form complete characters. Also, evaluating the robustness of language generation models on real-world input noises may require additional benchmarks beyond those used in this paper. We leave such exploration as an interesting direction for future work. 

Another limitation is that our method is designed specifically for the Chinese language. While we hypothesize that our method can also bring benefits to other languages with ideographic symbols, such as Kanji in Japanese, we leave such investigation to future work.

\section*{Broader Impact}

We expect our work to have a positive impact on the society. Firstly, we addressed the practical problem of handling input with real-world noises. Such noisy settings are very common in real-life applications. Our method, along with the evaluation framework, can help make language technologies more robust and reliable in real-world applications, especially for Chinese users. Secondly, we addressed the efficiency concern of large language models by significantly reducing both and training and inference time. This not only reduces the latency of these models in real-world applications, more importantly, helps reduce the environmental costs of using these large language models, moving further towards Green AI. All of our code and models are released with proper documentation in order to better facilitate the adoption of our work in a wide range of research and industrial applications. 

\section*{Acknowledgement}

This work is supported by the National Key Research and Development Program of China (No. 2020AAA0106500) and the National Natural Science Foundation of China (NSFC No. 62236004).

We thank Xu Han, Yusheng Su, Tianyu Gao and other members of THUNLP for their helpful discussion in the early stages of this work. We thank Jordan Boyd-Graber, Chen Zhao, Shi Feng, Neha Srikanth, Tonia Bleam, Leslie Li, and other members of UMD CLIP and Language Science Center for their helpful discussion and feedback. We also thank Nelson Liu and Canwen Xu for their constructive feedback on our early drafts. 
We especially appreciate the constructive reviews from TACL reviewers and action editors. 

This work is supported by the National Key Research and Development Program of China (No. 2020AAA0106500) and the National Natural Science Foundation of China (NSFC No. 62236004).

We thank Xu Han, Yusheng Su, Tianyu Gao and other members of THUNLP for their helpful discussion in the early stages of this work. We thank Jordan Boyd-Graber, Chen Zhao, Shi Feng, Neha Srikanth, Tonia Bleam, Leslie Li, and other members of UMD CLIP and Language Science Center for their helpful discussion and feedback. We also thank Nelson Liu and Canwen Xu for their constructive feedback on our early drafts. 
We especially appreciate the constructive reviews from TACL reviewers and action editors. 

\paragraph{Author contributions} Chenglei Si, Zhengyan Zhang, and Yingfa Chen wrote the code and conducted the experiments. Chenglei was in charge of tokenzer training and pretraining experiments, Zhengyan did the CWS experiments, Yingfa did the finetuning experiments. All three of them contributed to the analysis experiments.
Chenglei Si, Zhengyan Zhang, and Yingfa Chen wrote the initial draft; Fanchao Qi, Xiaozhi Wang, and Zhiyuan Liu significantly edited and improved the paper.
Yasheng Wang, Qun Liu, and Maosong Sun provided valuable advice to the research. Chenglei started this work back when he was visiting the THUNLP group in 2021. 

\bibliography{tacl2018}

\begin{thebibliography}{60}
\expandafter\ifx\csname natexlab\endcsname\relax\def\natexlab#1{#1}\fi

\bibitem[{Bender(2019)}]{BenderRule}
Emily Bender. 2019.
\newblock \href
  {https://thegradient.pub/the-benderrule-on-naming-the-languages-we-study-and-why-it-matters/}
  {{The \#BenderRule: On Naming the Languages We Study and Why It Matters}}.
\newblock \emph{The Gradient}.

\bibitem[{Bharadwaj et~al.(2016)Bharadwaj, Mortensen, Dyer, and
  Carbonell}]{bharadwaj-etal-2016-phonologically}
Akash Bharadwaj, David Mortensen, Chris Dyer, and Jaime Carbonell. 2016.
\newblock \href {https://aclanthology.org/D16-1153} {Phonologically aware
  neural model for named entity recognition in low resource transfer settings}.
\newblock In \emph{Proceedings of EMNLP}, pages 1462--1472.

\bibitem[{Cao et~al.(2018)Cao, Lu, Zhou, and Li}]{cw2vec}
Shaosheng Cao, Wei Lu, Jun Zhou, and Xiaolong Li. 2018.
\newblock \href
  {https://www.aaai.org/ocs/index.php/AAAI/AAAI18/paper/view/17444} {{cw2vec:
  Learning Chinese Word Embeddings with Stroke n-gram Information}}.
\newblock In \emph{Proceedings of AAAI}.

\bibitem[{Chang et~al.(2008)Chang, Galley, and Manning}]{CWS-NMT}
Pi-Chuan Chang, Michel Galley, and Christopher~D. Manning. 2008.
\newblock \href {https://aclanthology.org/W08-0336} {Optimizing {C}hinese
  {W}ord {S}egmentation for {M}achine {T}ranslation {P}erformance}.
\newblock In \emph{Proceedings of the Third Workshop on Statistical Machine
  Translation}.

\bibitem[{Chaudhary et~al.(2018)Chaudhary, Zhou, Levin, Neubig, Mortensen, and
  Carbonell}]{chaudhary-etal-2018-adapting}
Aditi Chaudhary, Chunting Zhou, Lori Levin, Graham Neubig, David~R. Mortensen,
  and Jaime Carbonell. 2018.
\newblock \href {https://aclanthology.org/D18-1366} {Adapting word embeddings
  to new languages with morphological and phonological subword
  representations}.
\newblock In \emph{Proceedings of EMNLP}, pages 3285--3295.

\bibitem[{Chen et~al.(2018)Chen, Chen, Liu, Yang, Lu, and Tang}]{BQ}
Jing Chen, Qingcai Chen, Xin Liu, Haijun Yang, Daohe Lu, and Buzhou Tang. 2018.
\newblock \href {https://doi.org/10.18653/v1/D18-1536} {{The BQ Corpus: A
  Large-scale Domain-specific {C}hinese Corpus For Sentence Semantic
  Equivalence Identification}}.
\newblock In \emph{Proceedings of EMNLP}.

\bibitem[{Clark et~al.(2021)Clark, Garrette, Turc, and Wieting}]{CANINE}
Jonathan Clark, Dan Garrette, Iulia Turc, and John Wieting. 2021.
\newblock \href {https://arxiv.org/abs/2103.06874} {{CANINE: Pre-training an
  Efficient Tokenization-Free Encoder for Language Representation}}.
\newblock \emph{Transactions of the Association for Computational Linguistics},
  10:73--91.

\bibitem[{Clark et~al.(2020)Clark, Luong, Le, and Manning}]{ELECTRA}
Kevin Clark, Minh{-}Thang Luong, Quoc~V. Le, and Christopher~D. Manning. 2020.
\newblock \href {https://openreview.net/forum?id=r1xMH1BtvB} {{ELECTRA:
  Pre-training Text Encoders as Discriminators Rather Than Generators}}.
\newblock In \emph{Proceedings of ICLR}.

\bibitem[{Coulmas(1991)}]{writing_system}
Florian Coulmas. 1991.
\newblock \href
  {https://books.google.com/books/about/Writing_Systems_of_the_World.html?id=VOywmavmZ3UC}
  {{The Writing Systems of the World}}.
\newblock pages 108--109. Blackwell Publishers.

\bibitem[{Cui et~al.(2020)Cui, Che, Liu, Qin, Wang, and Hu}]{MacBERT}
Yiming Cui, Wanxiang Che, Ting Liu, Bing Qin, Shijin Wang, and Guoping Hu.
  2020.
\newblock \href {https://doi.org/10.18653/v1/2020.findings-emnlp.58}
  {{Revisiting Pre-Trained Models for {C}hinese Natural Language Processing}}.
\newblock In \emph{Findings of EMNLP}.

\bibitem[{Cui et~al.(2019{\natexlab{a}})Cui, Che, Liu, Qin, Yang, Wang, and
  Hu}]{WWM}
Yiming Cui, Wanxiang Che, Ting Liu, Bing Qin, Ziqing Yang, Shijin Wang, and
  Guoping Hu. 2019{\natexlab{a}}.
\newblock \href {https://arxiv.org/abs/1906.08101} {{Pre-Training with Whole
  Word Masking for Chinese BERT}}.
\newblock \emph{IEEE/ACM TASLP}, 29:3504--3514.

\bibitem[{Cui et~al.(2019{\natexlab{b}})Cui, Liu, Che, Xiao, Chen, Ma, Wang,
  and Hu}]{CMRC}
Yiming Cui, Ting Liu, Wanxiang Che, Li~Xiao, Zhipeng Chen, Wentao Ma, Shijin
  Wang, and Guoping Hu. 2019{\natexlab{b}}.
\newblock \href {https://doi.org/10.18653/v1/D19-1600} {{A Span-Extraction
  Dataset for {C}hinese Machine Reading Comprehension}}.
\newblock In \emph{Proceedings of EMNLP-IJCNLP}.

\bibitem[{Devlin et~al.(2019)Devlin, Chang, Lee, and Toutanova}]{BERT}
Jacob Devlin, Ming-Wei Chang, Kenton Lee, and Kristina Toutanova. 2019.
\newblock \href {https://doi.org/10.18653/v1/N19-1423} {{BERT: Pre-training of
  Deep Bidirectional Transformers for Language Understanding}}.
\newblock In \emph{Proceedings of NAACL-HLT}.

\bibitem[{Diao et~al.(2020)Diao, Bai, Song, Zhang, and Wang}]{ZEN}
Shizhe Diao, Jiaxin Bai, Yan Song, Tong Zhang, and Yonggang Wang. 2020.
\newblock \href {https://doi.org/10.18653/v1/2020.findings-emnlp.425} {{ZEN:
  Pre-training {C}hinese Text Encoder Enhanced by N-gram Representations}}.
\newblock In \emph{Findings of EMNLP}.

\bibitem[{Dong et~al.(2016)Dong, Zhang, Zong, Hattori, and
  Di}]{dong2016character}
Chuanhai Dong, Jiajun Zhang, Chengqing Zong, Masanori Hattori, and Hui Di.
  2016.
\newblock \href
  {https://link.springer.com/chapter/10.1007/978-3-319-50496-4_20}
  {Character-based lstm-crf with radical-level features for chinese named
  entity recognition}.
\newblock In \emph{Natural Language Understanding and Intelligent
  Applications}, pages 239--250.

\bibitem[{Hao and Yang(2021)}]{phonology_new}
Yen-Chen Hao and Chung-Lin~Martin Yang. 2021.
\newblock \href
  {https://www.cambridge.org/core/journals/applied-psycholinguistics/article/abs/effect-of-secondlanguage-orthographic-input-on-the-phonological-encoding-of-mandarin-words/2EBEFD0811B23383FA496616F598566D}
  {{The effect of second-language orthographic input on the phonological
  encoding of Mandarin words}}.
\newblock \emph{Applied Psycholinguistics}.

\bibitem[{He et~al.(2021)He, Liu, Gao, and Chen}]{DeBERTa}
Pengcheng He, Xiaodong Liu, Jianfeng Gao, and Weizhu Chen. 2021.
\newblock \href {https://openreview.net/forum?id=r1xMH1BtvB} {{DeBERTa:
  Decoding-enhanced BERT with Disentangled Attention}}.
\newblock In \emph{Proceedings of ICLR}.

\bibitem[{Hu et~al.(2020)Hu, Richardson, Xu, Li, K{\"u}bler, and Moss}]{OCNLI}
Hai Hu, Kyle Richardson, Liang Xu, Lu~Li, Sandra K{\"u}bler, and Lawrence Moss.
  2020.
\newblock \href {https://doi.org/10.18653/v1/2020.findings-emnlp.314} {{OCNLI}:
  {O}riginal {C}hinese {N}atural {L}anguage {I}nference}.
\newblock In \emph{Findings of EMNLP}.

\bibitem[{Huang et~al.(2021)Huang, Li, Jiang, Zhang, Chen, Wang, and
  Xiao}]{huang-etal-2021-phmospell}
Li~Huang, Junjie Li, Weiwei Jiang, Zhiyu Zhang, Minchuan Chen, Shaojun Wang,
  and Jing Xiao. 2021.
\newblock \href {https://aclanthology.org/2021.acl-long.464} {{PHMOS}pell:
  Phonological and morphological knowledge guided {C}hinese spelling check}.
\newblock In \emph{Proceedings of ACL}, pages 5958--5967.

\bibitem[{Krell et~al.(2021)Krell, Kosec, Perez, and
  Fitzgibbon}]{kosec2021packing}
Mario~Michael Krell, Matej Kosec, Sergio~P. Perez, and Andrew Fitzgibbon. 2021.
\newblock \href {https://arxiv.org/abs/2107.02027} {{Efficient Sequence Packing
  without Cross-contamination: Accelerating Large Language Models without
  Impacting Performance}}.
\newblock \emph{arXiv preprint}, abs/2107.02027.

\bibitem[{Kudo(2018)}]{Kudo2018}
Taku Kudo. 2018.
\newblock \href {https://doi.org/10.18653/v1/P18-1007} {{Subword
  Regularization: Improving Neural Network Translation Models with Multiple
  Subword Candidates}}.
\newblock In \emph{Proceedings of ACL}.

\bibitem[{Kudo and Richardson(2018)}]{sentencepiece}
Taku Kudo and John Richardson. 2018.
\newblock \href {https://doi.org/10.18653/v1/D18-2012} {{S}entence{P}iece: A
  simple and language independent subword tokenizer and detokenizer for
  {N}eural {T}ext {P}rocessing}.
\newblock In \emph{Proceedings of EMNLP System Demonstrations}.

\bibitem[{Lai et~al.(2021)Lai, Liu, Feng, Huang, and Zhao}]{Lattice-BERT}
Yuxuan Lai, Yijia Liu, Yansong Feng, Songfang Huang, and Dongyan Zhao. 2021.
\newblock \href {https://doi.org/10.18653/v1/2021.naacl-main.137}
  {Lattice-{BERT}: {L}everaging {M}ulti-{G}ranularity {R}epresentations in
  {C}hinese {P}re-trained {L}anguage {M}odels}.
\newblock In \emph{Proceedings of NAACL-HLT}.

\bibitem[{Lan et~al.(2020)Lan, Chen, Goodman, Gimpel, Sharma, and
  Soricut}]{ALBERT}
Zhenzhong Lan, Mingda Chen, Sebastian Goodman, Kevin Gimpel, Piyush Sharma, and
  Radu Soricut. 2020.
\newblock \href {https://openreview.net/forum?id=H1eA7AEtvS} {{ALBERT:} {A}
  {L}ite {BERT} for {S}elf-supervised {L}earning of {L}anguage
  {R}epresentations}.
\newblock In \emph{Proceedings of ICLR}.

\bibitem[{Levesque et~al.(2012)Levesque, Davis, and
  Morgenstern}]{levesque2012winograd}
Hector Levesque, Ernest Davis, and Leora Morgenstern. 2012.
\newblock \href {http://commonsensereasoning.org/2011/papers/Levesque.pdf}
  {{The Winograd Schema Challenge}}.
\newblock In \emph{Thirteenth international conference on the principles of
  knowledge representation and reasoning}.

\bibitem[{Li and Sun(2007)}]{THUCNEWS}
Jingyang Li and Maosong Sun. 2007.
\newblock \href {https://aclanthology.org/D07-1081} {{Scalable Term Selection
  for Text Categorization}}.
\newblock In \emph{Proceedings of EMNLP}.

\bibitem[{Li et~al.(2019)Li, Meng, Sun, Han, Yuan, and Li}]{CWSnecessary}
Xiaoya Li, Yuxian Meng, Xiaofei Sun, Qinghong Han, Arianna Yuan, and Jiwei Li.
  2019.
\newblock \href {https://doi.org/10.18653/v1/P19-1314} {{Is Word Segmentation
  Necessary for Deep Learning of {C}hinese Representations?}}
\newblock In \emph{Proceedings of ACL}.

\bibitem[{Li and Sun(2009)}]{THULAC}
Zhongguo Li and Maosong Sun. 2009.
\newblock \href {https://doi.org/10.1162/coli.2009.35.4.35403} {{Punctuation as
  Implicit Annotations for {C}hinese Word Segmentation}}.
\newblock \emph{Computational Linguistics}.

\bibitem[{Liu et~al.(2017)Liu, Lu, Lo, and Neubig}]{liu-etal-2017-learning}
Frederick Liu, Han Lu, Chieh Lo, and Graham Neubig. 2017.
\newblock \href {https://aclanthology.org/P17-1188} {Learning character-level
  compositionality with visual features}.
\newblock In \emph{Proceedings of ACL}, pages 2059--2068.

\bibitem[{Liu et~al.(2019)Liu, Ott, Goyal, Du, Joshi, Chen, Levy, Lewis,
  Zettlemoyer, and Stoyanov}]{RoBERTa}
Yinhan Liu, Myle Ott, Naman Goyal, Jingfei Du, Mandar Joshi, Danqi Chen, Omer
  Levy, Mike Lewis, Luke Zettlemoyer, and Veselin Stoyanov. 2019.
\newblock \href {https://arxiv.org/abs/1907.11692} {{RoBERTa: A Robustly
  Optimized BERT Pretraining Approach}}.
\newblock \emph{arXiv preprint}, abs/1907.11692.

\bibitem[{Ma et~al.(2020)Ma, Cui, Si, Liu, Wang, and Hu}]{CharBERT}
Wentao Ma, Yiming Cui, Chenglei Si, Ting Liu, Shijin Wang, and Guoping Hu.
  2020.
\newblock \href {https://doi.org/10.18653/v1/2020.coling-main.4} {{C}har{BERT}:
  {C}haracter-aware {P}re-trained {L}anguage {M}odel}.
\newblock In \emph{Proceedings of COLING}.

\bibitem[{McCoy et~al.(2019)McCoy, Pavlick, and Linzen}]{McCoy2019RightFT}
R.~Thomas McCoy, Ellie Pavlick, and Tal Linzen. 2019.
\newblock \href {https://doi.org/10.18653/v1/p19-1334} {Right for the wrong
  reasons: Diagnosing syntactic heuristics in natural language inference}.
\newblock In \emph{Proceedings of ACL}.

\bibitem[{Meng et~al.(2019)Meng, Wu, Wang, Li, Nie, Yin, Li, Han, Sun, and
  Li}]{glyce}
Yuxian Meng, Wei Wu, Fei Wang, Xiaoya Li, Ping Nie, Fan Yin, Muyu Li, Qinghong
  Han, Xiaofei Sun, and Jiwei Li. 2019.
\newblock \href
  {https://proceedings.neurips.cc/paper/2019/hash/452bf208bf901322968557227b8f6efe-Abstract.html}
  {{Glyce: Glyph-vectors for Chinese Character Representations}}.
\newblock In \emph{Proceedings of NeurIPS}.

\bibitem[{Mikolov et~al.(2013)Mikolov, Sutskever, Chen, Corrado, and
  Dean}]{mikolov2013distributed}
Tomas Mikolov, Ilya Sutskever, Kai Chen, Greg~S Corrado, and Jeff Dean. 2013.
\newblock \href {https://arxiv.org/abs/1310.4546} {Distributed representations
  of words and phrases and their compositionality}.
\newblock In \emph{Proceedings of NeurIPS}, volume~26.

\bibitem[{Nzeyimana and Rubungo(2022)}]{Nzeyimana2022KinyaBERTAM}
Antoine Nzeyimana and Andre~Niyongabo Rubungo. 2022.
\newblock \href {https://aclanthology.org/2022.acl-long.367} {Kinyabert: a
  morphology-aware kinyarwanda language model}.
\newblock In \emph{Proceedings of ACL}.

\bibitem[{Peters et~al.(2018)Peters, Neumann, Iyyer, Gardner, Clark, Lee, and
  Zettlemoyer}]{ELMo}
Matthew~E. Peters, Mark Neumann, Mohit Iyyer, Matt Gardner, Christopher Clark,
  Kenton Lee, and Luke Zettlemoyer. 2018.
\newblock \href {https://doi.org/10.18653/v1/N18-1202} {{Deep Contextualized
  Word Representations}}.
\newblock In \emph{Proceedings of NAACL-HLT}.

\bibitem[{Schuster and Nakajima(2012)}]{WordPiece}
Mike Schuster and Kaisuke Nakajima. 2012.
\newblock \href {https://ieeexplore.ieee.org/document/6289079} {{Japanese and
  Korean voice search}}.
\newblock In \emph{Proceedings of the IEEE International Conference on
  Acoustics, Speech and Signal Processing}.

\bibitem[{Sennrich et~al.(2016)Sennrich, Haddow, and Birch}]{BPE}
Rico Sennrich, Barry Haddow, and Alexandra Birch. 2016.
\newblock \href {https://doi.org/10.18653/v1/P16-1162} {{Neural Machine
  Translation of Rare Words with Subword Units}}.
\newblock In \emph{Proceedings of ACL}.

\bibitem[{Si et~al.(2023)Si, Zhang, Chen, Wang, Liu, and Sun}]{READIN}
Chenglei Si, Zhengyan Zhang, Yingfa Chen, Xiaozhi Wang, Zhiyuan Liu, and
  Maosong Sun. 2023.
\newblock {READIN: A Chinese Multi-Task Benchmark with Realistic and Diverse
  Input Noises}.
\newblock \emph{arXiv}.

\bibitem[{Sun et~al.(2020)Sun, Yu, Yu, and Cardie}]{C3}
Kai Sun, Dian Yu, Dong Yu, and Claire Cardie. 2020.
\newblock \href {https://doi.org/10.1162/tacl_a_00305} {{Investigating Prior
  Knowledge for Challenging {C}hinese Machine Reading Comprehension}}.
\newblock \emph{Transactions of the Association for Computational Linguistics}.

\bibitem[{Sun et~al.(2016)Sun, Chen, Zhang, Guo, and Liu}]{THULAC-repo}
Maosong Sun, Xinxiong Chen, Kaixu Zhang, Zhipeng Guo, and Zhiyuan Liu. 2016.
\newblock \href {https://github.com/thunlp/THULAC-Python} {{THULAC: An
  Efficient Lexical Analyzer for Chinese}}.
\newblock \emph{GitHub}.

\bibitem[{Sun et~al.(2014)Sun, Lin, Yang, Ji, and Wang}]{sun2014radical}
Yaming Sun, Lei Lin, Nan Yang, Zhenzhou Ji, and Xiaolong Wang. 2014.
\newblock \href {https://arxiv.org/abs/1404.4714} {Radical-enhanced chinese
  character embedding}.
\newblock In \emph{Proceedings of COLING}, pages 279--286.

\bibitem[{Sun et~al.(2019)Sun, Wang, Li, Feng, Chen, Zhang, Tian, Zhu, Tian,
  and Wu}]{ERNIE}
Yu~Sun, Shuohuan Wang, Yukun Li, Shikun Feng, Xuyi Chen, Han Zhang, Xin Tian,
  Danxiang Zhu, Hao Tian, and Hua Wu. 2019.
\newblock \href {https://arxiv.org/abs/1904.09223} {{ERNIE: Enhanced
  Representation through Knowledge Integration}}.
\newblock \emph{arXiv preprint}, abs/1904.09223.

\bibitem[{Sun et~al.(2021)Sun, Li, Sun, Meng, Ao, He, Wu, and
  Li}]{sun-etal-2021-chinesebert}
Zijun Sun, Xiaoya Li, Xiaofei Sun, Yuxian Meng, Xiang Ao, Qing He, Fei Wu, and
  Jiwei Li. 2021.
\newblock \href {https://aclanthology.org/2021.acl-long.161} {{C}hinese{BERT}:
  {C}hinese pretraining enhanced by glyph and {P}inyin information}.
\newblock In \emph{Proceedings of ACL}, pages 2065--2075.

\bibitem[{Tay et~al.(2022)Tay, Tran, Ruder, Gupta, Chung, Bahri, Qin,
  Baumgartner, Yu, and Metzler}]{Charformer}
Yi~Tay, Vinh Tran, Sebastian Ruder, Jai Gupta, Hyung~Won Chung, Dara Bahri,
  Zhen Qin, Simon Baumgartner, Cong Yu, and Donald Metzler. 2022.
\newblock \href {https://arxiv.org/abs/2106.12672} {{Charformer: Fast Character
  Transformers via Gradient-based Subword Tokenization}}.
\newblock In \emph{Proceedings of ICLR}.

\bibitem[{Vaswani et~al.(2017)Vaswani, Shazeer, Parmar, Uszkoreit, Jones,
  Gomez, Kaiser, and Polosukhin}]{Transformer}
Ashish Vaswani, Noam Shazeer, Niki Parmar, Jakob Uszkoreit, Llion Jones,
  Aidan~N. Gomez, Lukasz Kaiser, and Illia Polosukhin. 2017.
\newblock \href
  {https://proceedings.neurips.cc/paper/2017/hash/3f5ee243547dee91fbd053c1c4a845aa-Abstract.html}
  {{Attention is All you Need}}.
\newblock In \emph{Proceedings of NeurIPS}.

\bibitem[{Wei et~al.(2021)Wei, Liu, Guo, and Jiang}]{BBPE}
Junqiu Wei, Qun Liu, Yinpeng Guo, and Xin Jiang. 2021.
\newblock \href {https://arxiv.org/abs/2101.09469} {{Training Multilingual
  Pre-trained Language Model with Byte-leve}}.
\newblock \emph{arXiv preprint}, abs/2101.09469.

\bibitem[{Wei et~al.(2019)Wei, Ren, Li, Huang, Liao, Wang, Lin, Jiang, Chen,
  and Liu}]{NEZHA}
Junqiu Wei, Xiaozhe Ren, Xiaoguang Li, Wenyong Huang, Yi~Liao, Yasheng Wang,
  Jiashu Lin, Xin Jiang, Xiao Chen, and Qun Liu. 2019.
\newblock \href {http://arxiv.org/abs/1909.00204} {{NEZHA: Neural
  Contextualized Representation for Chinese Language Understanding}}.
\newblock \emph{arXiv}, abs/1904.00204.

\bibitem[{Xiao et~al.(2021)Xiao, Li, Zhang, Sun, Tian, Wu, and
  Wang}]{ERNIE-GRAM}
Dongling Xiao, Yu-Kun Li, Han Zhang, Yu~Sun, Hao Tian, Hua Wu, and Haifeng
  Wang. 2021.
\newblock \href {https://doi.org/10.18653/v1/2021.naacl-main.136} {{ERNIE-Gram:
  Pre-Training with Explicitly N-Gram Masked Language Modeling for Natural
  Language Understanding}}.
\newblock In \emph{Proceedings of NAACL-HLT}.

\bibitem[{Xu et~al.(2020{\natexlab{a}})Xu, Dong, Yu, Tian, Liu, Li, and
  Zhang}]{CLUENER}
Liang Xu, Qianqian Dong, Cong Yu, Yin Tian, Weitang Liu, Lu~Li, and Xuanwei
  Zhang. 2020{\natexlab{a}}.
\newblock \href {https://arxiv.org/abs/2001.04351} {{CLUENER2020: Fine-grained
  Name Entity Recognition for Chinese}}.
\newblock \emph{arXiv preprint}, abs/2001.04351.

\bibitem[{Xu et~al.(2020{\natexlab{b}})Xu, Hu, Zhang, Li, Cao, Li, Xu, Sun, Yu,
  Yu, Tian, Dong, Liu, Shi, Cui, Li, Zeng, Wang, Xie, Li, Patterson, Tian,
  Zhang, Zhou, Liu, Zhao, Zhao, Yue, Zhang, Yang, Richardson, and
  Lan}]{CLUEBenchmark}
Liang Xu, Hai Hu, Xuanwei Zhang, Lu~Li, Chenjie Cao, Yudong Li, Yechen Xu, Kai
  Sun, Dian Yu, Cong Yu, Yin Tian, Qianqian Dong, Weitang Liu, Bo~Shi, Yiming
  Cui, Junyi Li, Jun Zeng, Rongzhao Wang, Weijian Xie, Yanting Li, Yina
  Patterson, Zuoyu Tian, Yiwen Zhang, He~Zhou, Shaoweihua Liu, Zhe Zhao, Qipeng
  Zhao, Cong Yue, Xinrui Zhang, Zhengliang Yang, Kyle Richardson, and Zhenzhong
  Lan. 2020{\natexlab{b}}.
\newblock \href {https://doi.org/10.18653/v1/2020.coling-main.419} {{CLUE: A
  Chinese Language Understanding Evaluation Benchmark}}.
\newblock In \emph{Proceedings of COLING}.

\bibitem[{Xue et~al.(2022)Xue, Barua, Constant, Al-Rfou, Narang, Kale, Roberts,
  and Raffel}]{ByT5}
Linting Xue, Aditya Barua, Noah Constant, Rami Al-Rfou, Sharan Narang, Mihir
  Kale, Adam Roberts, and Colin Raffel. 2022.
\newblock \href {https://arxiv.org/abs/2105.13626} {{ByT5: Towards a token-free
  future with pre-trained byte-to-byte models}}.
\newblock \emph{Transactions of the Association for Computational Linguistics}.

\bibitem[{Yu et~al.(2017)Yu, Jian, Xin, and Song}]{yu2017joint}
Jinxing Yu, Xun Jian, Hao Xin, and Yangqiu Song. 2017.
\newblock \href {https://aclanthology.org/D17-1027/} {Joint embeddings of
  chinese words, characters, and fine-grained subcharacter components}.
\newblock In \emph{Proceedings of EMNLP}, pages 286--291.

\bibitem[{Zhang et~al.(2021{\natexlab{a}})Zhang, Li, and Li}]{AMBERT}
Xinsong Zhang, Pengshuai Li, and Hang Li. 2021{\natexlab{a}}.
\newblock \href {https://doi.org/10.18653/v1/2021.findings-acl.37} {{AMBERT: A
  Pre-trained Language Model with Multi-Grained Tokenization}}.
\newblock In \emph{Findings of ACL}.

\bibitem[{Zhang et~al.(2019{\natexlab{a}})Zhang, Baldridge, and
  He}]{Zhang2019PAWSPA}
Yuan Zhang, Jason Baldridge, and Luheng He. 2019{\natexlab{a}}.
\newblock \href {https://doi.org/10.18653/v1/n19-1131} {Paws: Paraphrase
  adversaries from word scrambling}.
\newblock In \emph{Proceedings of NAACL-HLT}.

\bibitem[{Zhang et~al.(2019{\natexlab{b}})Zhang, Liu, Zhu, Zheng, Liu, Wang,
  Chen, and Zhai}]{zhang2019learning}
Yun Zhang, Yongguo Liu, Jiajing Zhu, Ziqiang Zheng, Xiaofeng Liu, Weiguang
  Wang, Zijie Chen, and Shuangqing Zhai. 2019{\natexlab{b}}.
\newblock \href {https://dl.acm.org/doi/10.1145/3357384.3358005} {Learning
  chinese word embeddings from stroke, structure and pinyin of characters}.
\newblock In \emph{Proceedings of CIKM}, pages 1011--1020.

\bibitem[{Zhang et~al.(2021{\natexlab{b}})Zhang, Gu, Han, Chen, Xiao, Sun, Yao,
  Qi, Guan, Ke, Cai, Zeng, Tan, Liu, Huang, Han, Liu, Zhu, and Sun}]{CPM2}
Zhengyan Zhang, Yuxian Gu, Xu~Han, Shengqi Chen, Chaojun Xiao, Zhenbo Sun, Yuan
  Yao, Fanchao Qi, Jian Guan, Pei Ke, Yanzheng Cai, Guoyang Zeng, Zhixing Tan,
  Zhiyuan Liu, Minlie Huang, Wentao Han, Yang Liu, Xiaoyan Zhu, and Maosong
  Sun. 2021{\natexlab{b}}.
\newblock \href {https://arxiv.org/abs/2106.10715} {{CPM-2: Large-scale
  Cost-effective Pre-trained Language Models}}.
\newblock \emph{arXiv preprint}, abs/2106.10715.

\bibitem[{Zhang et~al.(2020)Zhang, Han, Zhou, Ke, Gu, Ye, Qin, Su, Ji, Guan,
  Qi, Wang, Zheng, Zeng, Cao, Chen, Li, Sun, Liu, Huang, Han, Tang, Li, Zhu,
  and Sun}]{CPM}
Zhengyan Zhang, Xu~Han, Hao Zhou, Pei Ke, Yuxian Gu, Deming Ye, Yujia Qin,
  Yusheng Su, Haozhe Ji, Jian Guan, Fanchao Qi, Xiaozhi Wang, Yanan Zheng,
  Guoyang Zeng, Huanqi Cao, Shengqi Chen, Daixuan Li, Zhenbo Sun, Zhiyuan Liu,
  Minlie Huang, Wentao Han, Jie Tang, Juanzi Li, Xiaoyan Zhu, and Maosong Sun.
  2020.
\newblock \href {https://arxiv.org/abs/2012.00413} {{CPM: A Large-scale
  Generative Chinese Pre-trained Language Model}}.
\newblock \emph{AI Open}.

\bibitem[{Zheng et~al.(2019)Zheng, Huang, and Sun}]{CHID}
Chujie Zheng, Minlie Huang, and Aixin Sun. 2019.
\newblock \href {https://doi.org/10.18653/v1/P19-1075} {{C}h{ID}: A
  {L}arge-scale {C}hinese {ID}iom {D}ataset for {C}loze {T}est}.
\newblock In \emph{Proceedings of ACL}.

\bibitem[{Zhu(2020)}]{MVP-BERT}
Wei Zhu. 2020.
\newblock \href {https://arxiv.org/abs/2011.08539} {{MVP-BERT: Redesigning
  Vocabularies for Chinese BERT and Multi-Vocab Pretraining}}.
\newblock \emph{arXiv preprint}, abs/2011.08539.

\end{thebibliography}
\bibliographystyle{acl_natbib}

\clearpage


\end{CJK*}

\end{document}